\documentclass[conference,onecolumn,12pt]{IEEEtran}
\IEEEoverridecommandlockouts
\usepackage{cite}
\usepackage{amsmath,amssymb,amsfonts}
\usepackage{bm}
\usepackage{algorithmic}
\usepackage{graphicx}
\graphicspath{{figs/}}
\usepackage{mdframed}
\usepackage{textcomp}
\usepackage[misc]{ifsym}
\usepackage{xcolor}
\def\BibTeX{{\rm B\kern-.05em{\sc i\kern-.025em b}\kern-.08em
    T\kern-.1667em\lower.7ex\hbox{E}\kern-.125emX}}

\usepackage{times}
\usepackage{soul}
\usepackage{url}
\usepackage[hidelinks]{hyperref}
\usepackage[utf8]{inputenc}
\usepackage[small]{caption}
\usepackage{amsthm}
\usepackage{transparent}
\usepackage{booktabs}
\usepackage{setspace}
\usepackage{multirow}
\usepackage{overpic}
\usepackage{tikz}
\usetikzlibrary{shapes}
\usepackage{pgfplots}
\pgfplotsset{compat=newest}
\usepackage{subcaption}
\usepackage{cleveref}

\newcommand{\iou}{\mathit{IoU}}
\newcommand{\auroc}{\mathit{AUROC}}
\newcommand{\ioubev}{\mathit{IoU}_{\!\mathrm{BEV}}}
\newcommand{\iouthreed}{\mathit{IoU}_{\!\mathrm{3D}}}
\newcommand{\map}{\mathit{mAP}}
\newcommand{\NMS}{\mathrm{NMS}}
\newcommand{\R}{\mathbb{R}}

\newcommand{\aptiv}{Aptiv}

\begin{document}

\title{LMD: Light-weight Prediction Quality Estimation for Object Detection in Lidar Point Clouds}


\author{Tobias Riedlinger${}^{1\ast}$, Marius Schubert${}^{1\ast}$, Sarina Penquitt${}^{1\ast}$, Jan-Marcel Kezmann${}^{1\ast}$,\\ Pascal Colling${}^2$, Karsten Kahl${}^1$, Lutz Roese-Koerner${}^2$, Michael Arnold${}^2$,\\ Urs Zimmermann${}^2$, Matthias Rottmann${}^1$ \\
\small ${}^1$School of Mathematics and Natural Sciences, IZMD, University of Wuppertal, Germany \\
\small\texttt{\{riedlinger@,schubert@math.,penquitt@,kkahl@math.,rottmann@math.\}uni-wuppertal.de,}\\\small\texttt{jankezmann@t-online.de}\\
${}^2$Aptiv Services Deutschland GmbH \\
\small\texttt{\{pascal.colling,lutz.roese-koerner,michael.arnold,urs.zimmermann\}@aptiv.com}
}

\newcommand{\PC}[1]{\textcolor{orange}{#1}}
\newcommand{\outPC}[1]{\textcolor{orange}{\sout{#1}}}
\newcommand{\MS}[1]{\textcolor{green}{#1}}
\newcommand{\outMS}[1]{\textcolor{green}{\sout{#1}}}
\newcommand{\TR}[1]{\textcolor{cyan}{#1}}
\newcommand{\outTR}[1]{\textcolor{cyan}{\st{#1}}}

\maketitle

\begin{abstract}
    Object detection on Lidar point cloud data is a promising technology for autonomous driving and robotics which has seen a significant rise in performance and accuracy during recent years.
    Particularly uncertainty estimation is a crucial component for down-stream tasks and deep neural networks remain error-prone even for predictions with high confidence. 
    Previously proposed methods for quantifying prediction uncertainty tend to alter the training scheme of the detector or rely on prediction sampling which results in vastly increased inference time.
    In order to address these two issues, we propose LidarMetaDetect (LMD), a light-weight post-processing scheme for prediction quality estimation.
    Our method can easily be added to any pre-trained Lidar object detector without altering anything about the base model and is purely based on post-processing, therefore, only leading to a negligible computational overhead.
    Our experiments show a significant increase of statistical reliability in separating true from false predictions.
    We propose and evaluate an additional application of our method leading to the detection of annotation errors.
    Explicit samples and a conservative count of annotation error proposals indicates the viability of our method for large-scale datasets like KITTI and nuScenes.
    On the widely-used nuScenes test dataset, 43 out of the top 100 proposals of our method indicate, in fact, erroneous annotations.
    
\end{abstract}


\def\thefootnote{*}\footnotetext{Equal contribution.}
\begin{figure*}
    \begin{mdframed}[backgroundcolor=black]
        \begin{overpic}[width=0.5\linewidth]{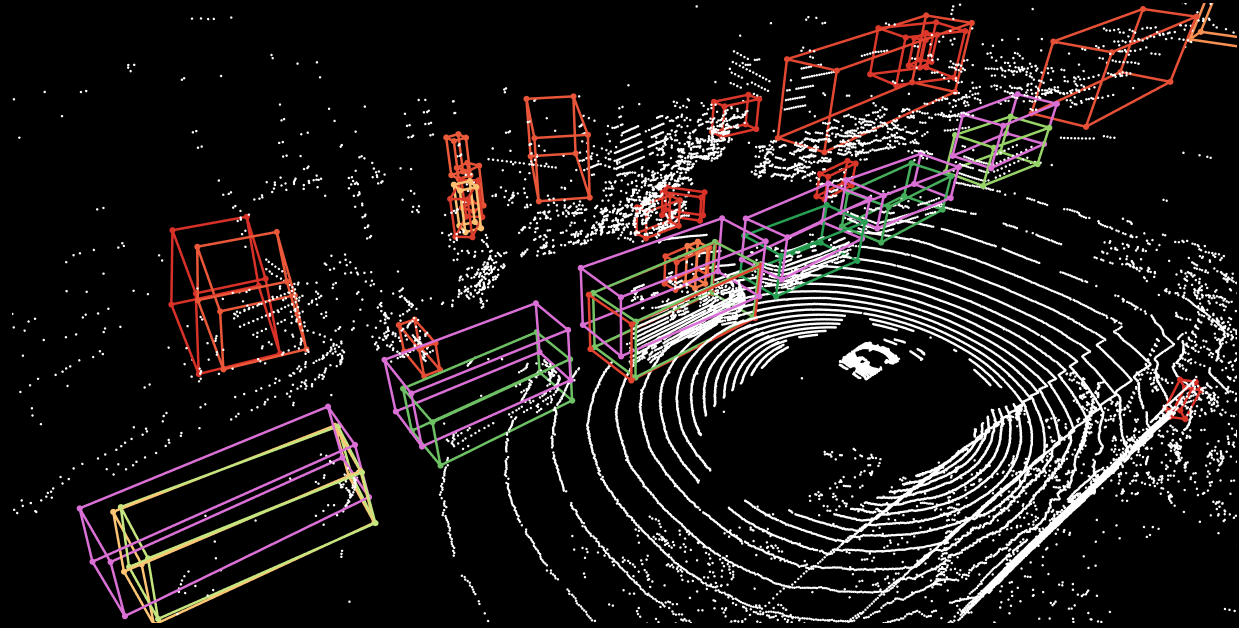}
            \put(1,47){\color{white}\footnotesize DNN objectness score}
        \end{overpic}%
		\begin{overpic}[width=0.5\linewidth,trim={0 0 0 0.8cm},clip]{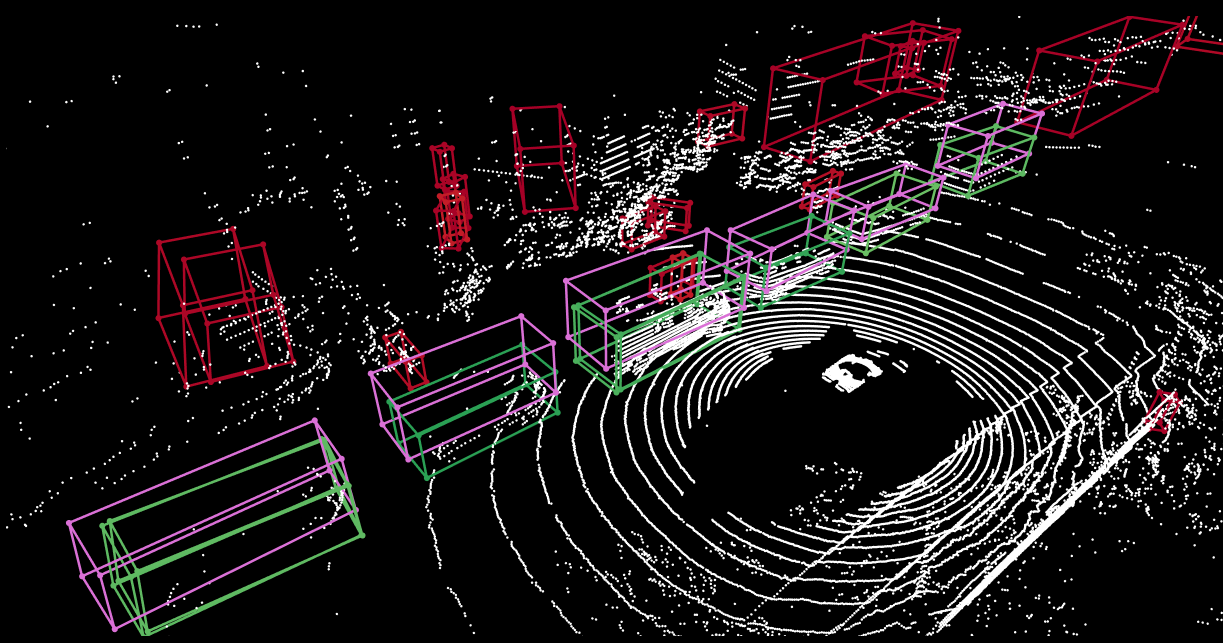}
            \put(1,47){\color{white}\footnotesize LMD score}
        \end{overpic}
		\includegraphics[width=0.25\linewidth]{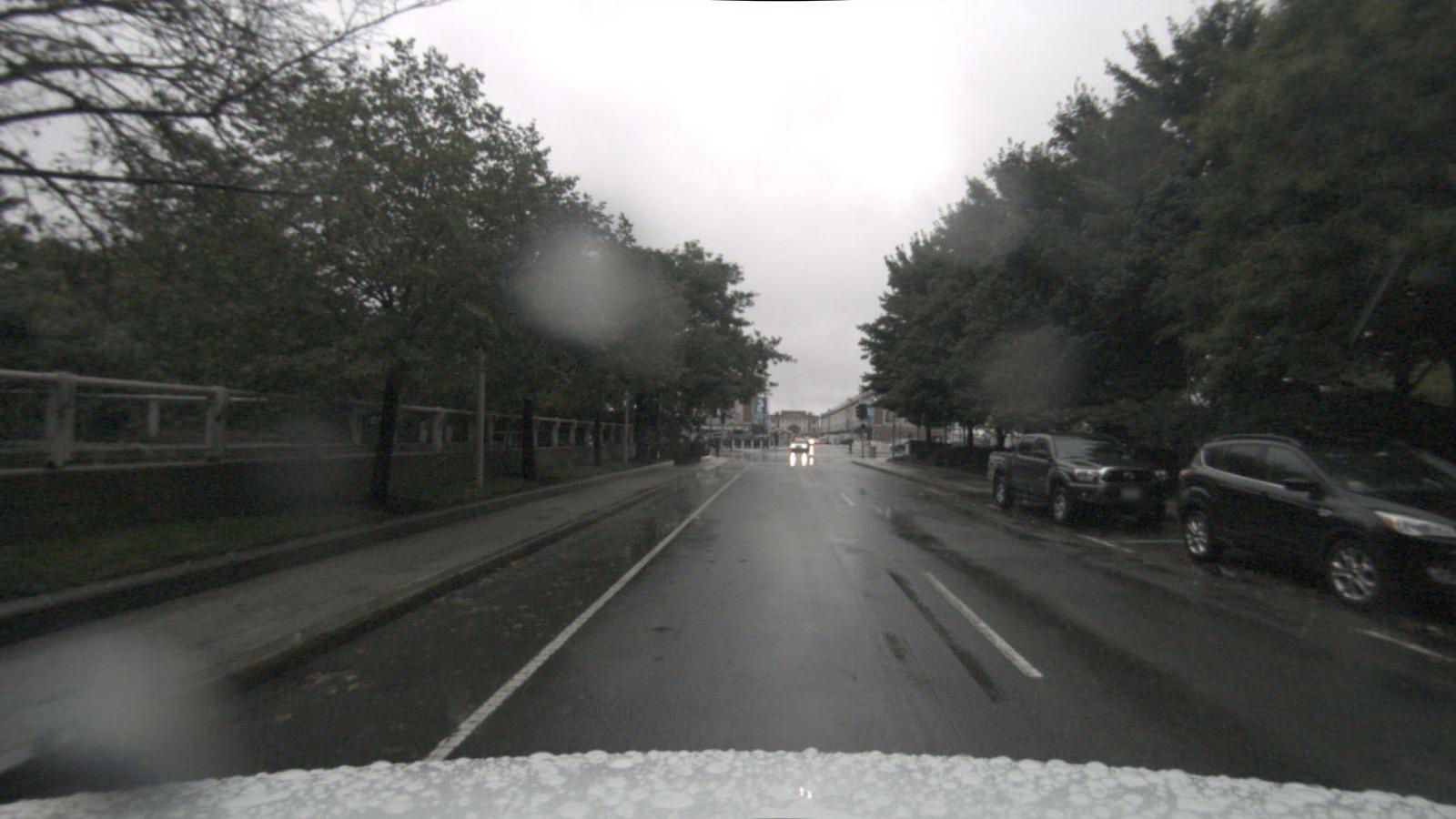}%
		\includegraphics[width=0.25\linewidth]{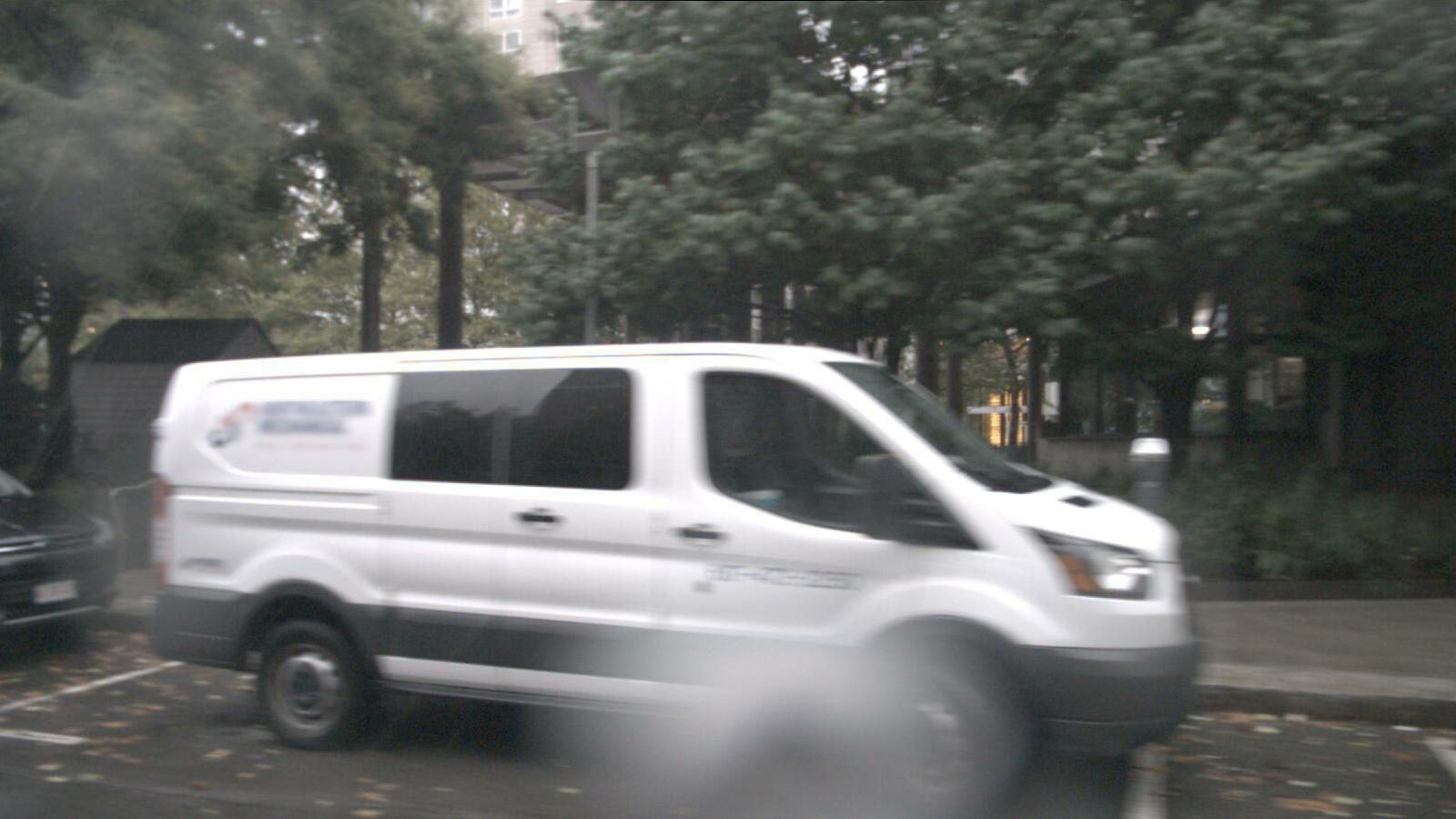}%
		\includegraphics[width=0.25\linewidth]{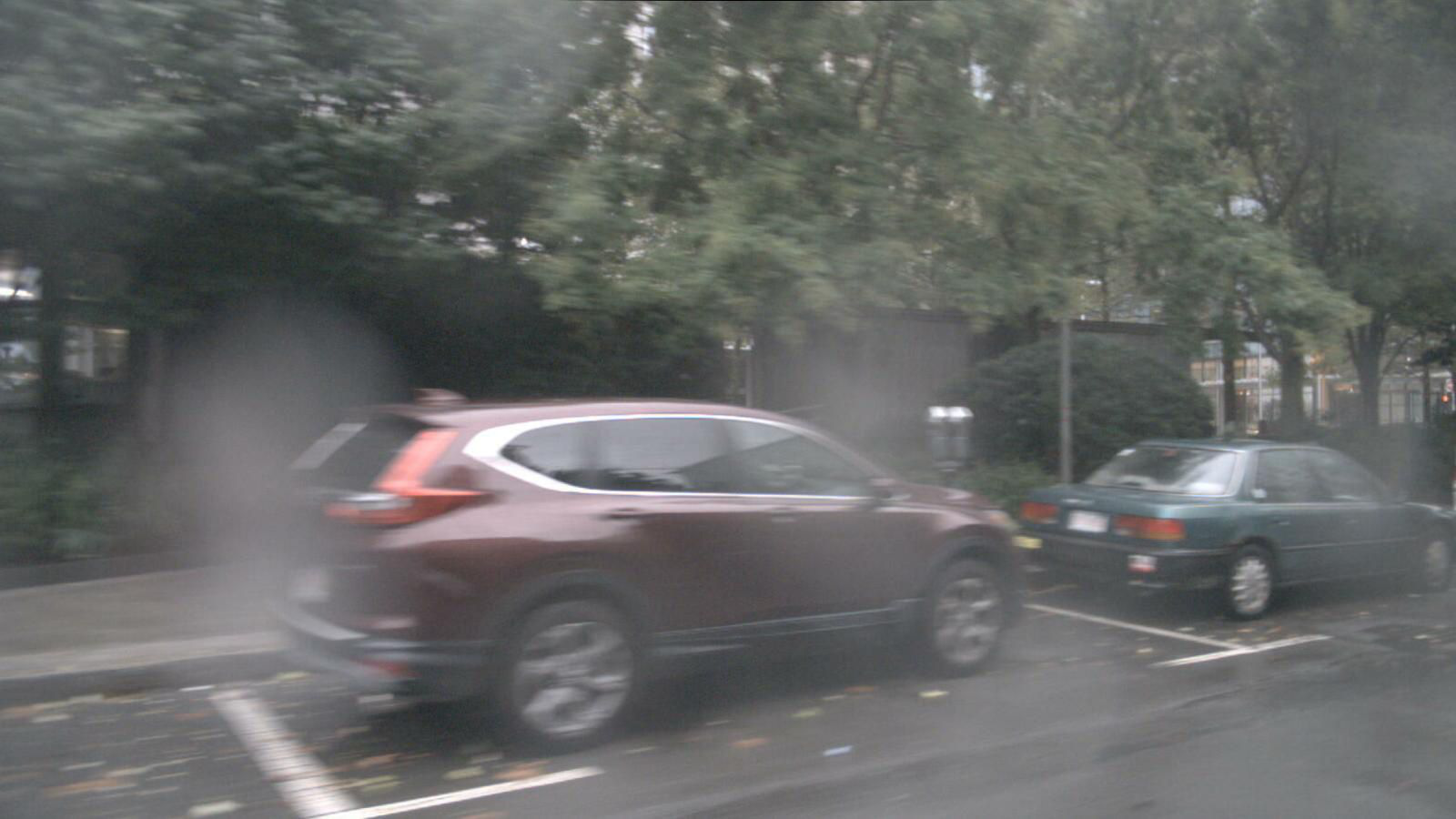}%
		\includegraphics[width=0.25\linewidth]{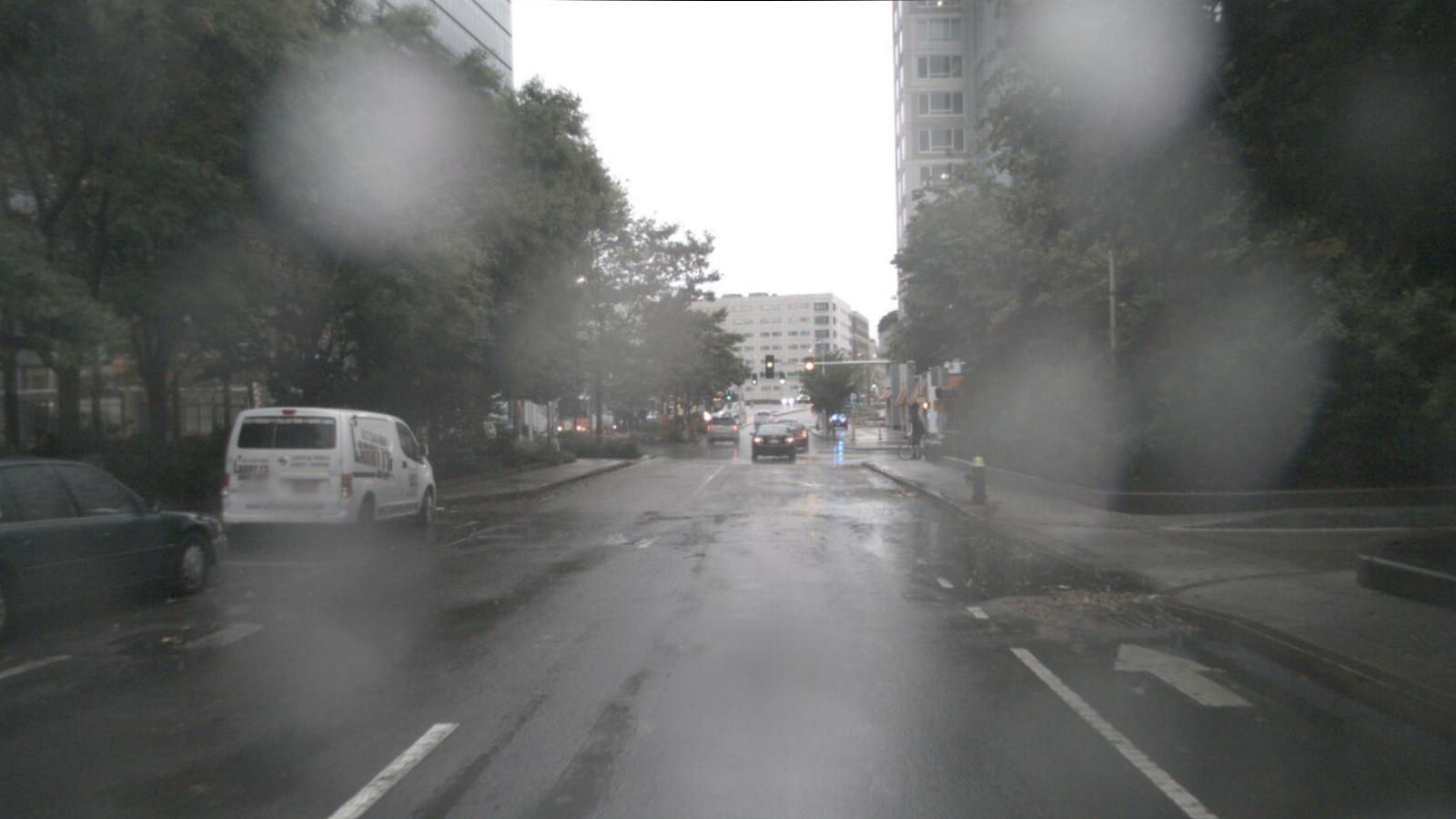}
    \end{mdframed}
	\caption{
		Prediction of a Lidar point cloud object detector with the native objectness score (\emph{left}) and LMD meta classifier scores (\emph{right}) and corresponding camera images below.
		Detections based on the objectness score are highly threshold-dependent and may lead to false positive detections.
		Detections based on LMD scores are more reliable and separate true from false predictions in a sharper way.
	}
\end{figure*}
\section{Introduction}


In recent years, deep learning has achieved great advances in the field of 3D object detection on Lidar data~\cite{lang_pointpillars_2019,yan_second_2018,yang_pixor_2018,yin_center-based_2021}. 
Deep neural network (DNN) architectures for this task are well-developed, however, there is little work in the area of uncertainty quantification (UQ) for such models~\cite{pitropov_lidar-mimo_2022,chen_monorun_2021,meyer_lasernet_2019,meyer_learning_2020,yang_uncertainty_2021}. 
UQ is crucial for deployment of DNN-based object detection in the real world, since DNNs as statistical models statistically make erroneous predictions.
Down-stream algorithms are supposed to further process the predictions of perception algorithms and rely on statistically accurate and meaningful UQ.
Aleatoric uncertainty is usually estimated by adding variance parameters to the network prediction and fitting them to data under a specific assumption for the distribution of residuals~\cite{chen_monorun_2021,meyer_learning_2020,meyer_lasernet_2019,feng_leveraging_2019,feng_towards_2018}.
Such approaches usually alter the training objective of the detector by appealing to the negative log-likelihood loss for normally distributed residuals.
Epistemic uncertainty is oftentimes estimated via Monte-Carlo (MC) dropout~\cite{chen_monorun_2021} or deep ensembles~\cite{yang_uncertainty_2021}.
In such approaches, model sampling leads to a significant increase in inference time.
Inspired by lines of research~\cite{schubert_metadetect_2021,riedlinger_gradient-based_2023} in the field of 2D object detection on camera images, we develop a framework for UQ in 3D object detection for Lidar point clouds.
This approach does not alter the training objective and can be applied to any pre-trained object detector and does not require prediction sampling.
Our framework, called LidarMetaDetect (short LMD), performs two UQ tasks: (1) meta classification, which aims at estimating the probability of a given prediction being a true positive vs.\ being a false positive; (2) meta regression, which estimates the localization quality of a prediction compared with the ground truth. 
Note that, outside of the context of UQ for DNNs, the terms meta classification and meta regression refer to different concepts, see~\cite{lin_meta-classification_2003} and~\cite{stanley_meta-regression_2005}, respectively.
LMD operates as a post-processing module and can be combined with any DNN without modifying it. 
Our methods learn on a small sample of data to assess the DNN's reliability in a frequentist sense at runtime, i.e.,  in the absence of ground truth. 
In essence, we handcraft a number of uncertainty scores on bounding box level, by which we convert both UQ tasks into structured machine learning tasks. 
To the best of our knowledge, our method is the first purely post-processing-based UQ method for 3D object detection based on Lidar point clouds. 
We conduct in-depth numerical studies on the KITTI~\cite{geiger_are_2012}, nuScenes~\cite{caesar_nuscenes_2020} as well as a propriety dataset.
including comparisons of our methods with baseline methods on common uncertainty quantification benchmarks, ablation studies of relevant parameters and the relevance of our uncertainty features. 
This is complemented with down stream tasks where (1) we demonstrate that our UQ increases the separation of true and false predictions and leads to well-calibrated confidence estimates and (2) we show that our UQ can be utilized for the detection of erroneous annotations in Lidar object detection datasets. 
We evaluate our method's annotations error detection capabilities by reviewing its proposals on moderate samples from KITTI and nuScenes.
Our contributions can be summarized as follows:

\begin{itemize}
    \item We develop the first purely post-processing based UQ framework for 3D object detection in Lidar point clouds.
    \item We compare our UQ methods to baselines and show that they clearly outperform the DNN's built-in estimates of reliability.
    \item We find annotation errors in the most commonly used publicly available Lidar object detection datasets, i.e., KITTI and nuScenes. 
\end{itemize}
We make our code publicly available at \url{https://github.com/JanMarcelKezmann/MetaDetect3D}.

\section{Related Work}
In recent years, technologically sophisticated methods such as perception in Lidar point clouds have received attention in the UQ branch due to their potential industrial relevance in the autonomous driving sector.
Methods for 3D object detection roughly fall into the categories of aleatoric and epistemic UQ.
Aleatoric UQ methods usually build on estimating distributional noise by adding a variance output for each regression variable while epistemic UQ methods utilize some kind of model sampling either appealing to MC dropout or deep ensembles.
Meyer et al.~\cite{meyer_lasernet_2019} estimate aleatoric uncertainty by a two-dimensional discretization scheme over the Lidar range and introducing a variance-weighted regression loss for a multi-modal distributional prediction in order to improve detection performance.
Meyer and Thakurdesai~\cite{meyer_learning_2020} estimate aleatoric uncertainty by adding scale regression variables to the network output, modeling Laplace-distributed residuals under a label noise assumption via a Kullback-Leibler divergence loss.
Feng et al.~\cite{feng_leveraging_2019} estimate heteroscedastic aleatoric uncertainty for the region proposal and the detection head of an object detector separately by modeling diagonal-covariance normally distributed bounding box regression.
Feng et al.~\cite{feng_towards_2018} achieve joint estimation of aleatoric and epistemic UQ by adding regression variables that model the covariance diagonal of a multi-variate normal distribution of the four bounding box parameters alongside MC dropout total variance for the epistemic component.
Chen et al.~\cite{chen_monorun_2021} extract aleatoric uncertainty information from a self-supervised projection-reconstruction mechanism propagated to 3D object detection on camera images.
Further, epistemic uncertainty of object localization is quantified via MC dropout.
Yang et al.~\cite{yang_uncertainty_2021} perform UQ for 3D object detection on Lidar and extend the multi-input multi-output model MIMO~\cite{havasi_training_2021} which modifies the network to be supplied simultaneously with $n$ inputs and providing $n$ outputs.
This simulates a deep ensemble at inference time at the cost of increased memory consumption for input and output layers.

In the field of 2D object detection in camera images by DNNs, methods for UQ have been developed in a series of works~\cite{schubert_metadetect_2021,riedlinger_gradient-based_2023} related with research on UQ in semantic segmentation~\cite{rottmann_detection_2020,rottmann_prediction_2020}.
Schubert et al.~\cite{schubert_metadetect_2021} utilize the pre-NMS anchor statistics in a post-processing approach to obtain box-wise confidence and $\iou$-estimates.
Riedlinger et al.~\cite{riedlinger_gradient-based_2023} use instance-wise gradient scores in a post-processing scheme to obtain calibrated uncertainty estimates improving detection performance.
Inspired by these lines of research, we develop a framework for UQ in 3D object detection for Lidar point clouds.
We use lightweight post-processing models on top of a pre-trained Lidar point cloud object detector in order to obtain improved uncertainty and $\iou$-estimates.
In contrast to previous work, our approach has the advantage that it may be applied to any pre-trained object detector without alteration of training or architecture and does not carry the computational and memory cost of sampling weights in a Bayesian manner like MC dropout or deep ensembles.
We show that this approach leads to more reliable object detection predictions and that it can be applied in an intuitive way in order to detect annotation errors in object detection datasets. 

\section{Proposed Method}\label{sec:method}
In this section we describe our post-processing mechanism and how it can be applied to improve detection performance and to detect annotation errors.
Our method assumes an object detector $f(\cdot)$ which maps point clouds $\bm{X}$ to a list of $N$ bounding boxes
\begin{equation}
    f(\bm{X}) = \left\{\widehat{b}^1, \ldots, \widehat{b}^{N}\right\}.
\end{equation}
Point clouds $\bm{X} = (\bm{p}_1, \ldots, \bm{p}_{N_\mathrm{pt}})$ consist of Lidar points $\bm{p} = (x, y, z, r) \in \R^4$ represented by three coordinates $(x, y, z)$ and a reflectance value $r$ each.
Bounding boxes are represented by features \(\widehat{b}^j (\bm{X}) = ( \widehat{x}^j, \widehat{y}^j, \widehat{z}^j, \widehat{\ell}^j, \widehat{w}^j, \widehat{h}^j, \widehat{\theta}^j, \widehat{s}^j, \widehat{\pi}_1^j, \ldots, \widehat{\pi}_C^j )\).
Here, $\widehat{x}^j, \widehat{y}^j, \widehat{z}^j, \widehat{\ell}^j, \widehat{w}^j, \widehat{h}^j, \widehat{\theta}^j$ define the bounding box geometry, $\widehat{s}^j$ is the objectness score and $(\widehat{\pi}_1^j, \ldots, \widehat{\pi}_C^j)$ is the predicted categorical probability distribution.
The latter defines the predicted class $\widehat{\kappa}^j = \mathrm{argmax}_{c = 1, \ldots, C} \, \widehat{\pi}_c^j$ while the objectness score $\widehat{s}^j$ is the model's native confidence estimate for each prediction.
Out of the $N$ bounding boxes, only a small amount $N_\NMS$ will be left after non-maximum suppression (NMS) filtering and contribute to the final prediction of the detector
\begin{equation}
    \NMS[f(\bm{X})] = \left\{ \widehat{b}^{i} :\, i \in I_\NMS \right\},
\end{equation}
where we let $I_\NMS \subset \{1, \ldots, N\}$ denote the post-NMS index set indicating survivor boxes.
\begin{figure}[t]
    \centering
    \resizebox{0.4\linewidth}{!}{
    \small
\begingroup%
  \makeatletter%
  \providecommand\color[2][]{%
    \errmessage{(Inkscape) Color is used for the text in Inkscape, but the package 'color.sty' is not loaded}%
    \renewcommand\color[2][]{}%
  }%
  \providecommand\transparent[1]{%
    \errmessage{(Inkscape) Transparency is used (non-zero) for the text in Inkscape, but the package 'transparent.sty' is not loaded}%
    \renewcommand\transparent[1]{}%
  }%
  \providecommand\rotatebox[2]{#2}%
  \newcommand*\fsize{\dimexpr\f@size pt\relax}%
  \newcommand*\lineheight[1]{\fontsize{\fsize}{#1\fsize}\selectfont}%
  \ifx\svgwidth\undefined%
    \setlength{\unitlength}{192.33415032bp}%
    \ifx\svgscale\undefined%
      \relax%
    \else%
      \setlength{\unitlength}{\unitlength * \real{\svgscale}}%
    \fi%
  \else%
    \setlength{\unitlength}{\svgwidth}%
  \fi%
  \global\let\svgwidth\undefined%
  \global\let\svgscale\undefined%
  \makeatother%
  \begin{picture}(1,1.03513502)%
    \lineheight{1}%
    \setlength\tabcolsep{0pt}%
    \put(0,0){\includegraphics[width=\unitlength,page=1]{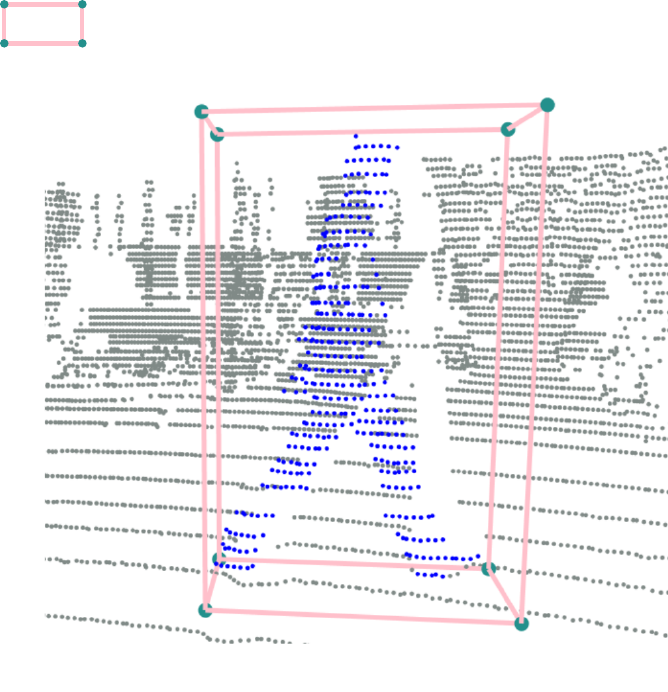}}%
    \put(0.15825976,0.9800502){\color[rgb]{0,0,0}\makebox(0,0)[lt]{\lineheight{1.25}\smash{\begin{tabular}[t]{l}$\widehat{b}^i$\end{tabular}}}}%
    \put(0,0){\includegraphics[width=\unitlength,page=2]{../figs/point_counts.pdf}}%
    \put(0.15825976,0.89061901){\color[rgb]{0,0,0}\makebox(0,0)[lt]{\lineheight{1.25}\smash{\begin{tabular}[t]{l}$\bm{X} \cap \widehat{b}^i$\end{tabular}}}}%
    \put(0.15825976,0.80690294){\color[rgb]{0,0,0}\makebox(0,0)[lt]{\lineheight{1.25}\smash{\begin{tabular}[t]{l}$\bm{X} \setminus \widehat{b}^i$\end{tabular}}}}%
  \end{picture}%
\endgroup%

    }%
    \resizebox{0.4\linewidth}{!}{
    \small
\begingroup%
  \makeatletter%
  \providecommand\color[2][]{%
    \errmessage{(Inkscape) Color is used for the text in Inkscape, but the package 'color.sty' is not loaded}%
    \renewcommand\color[2][]{}%
  }%
  \providecommand\transparent[1]{%
    \errmessage{(Inkscape) Transparency is used (non-zero) for the text in Inkscape, but the package 'transparent.sty' is not loaded}%
    \renewcommand\transparent[1]{}%
  }%
  \providecommand\rotatebox[2]{#2}%
  \newcommand*\fsize{\dimexpr\f@size pt\relax}%
  \newcommand*\lineheight[1]{\fontsize{\fsize}{#1\fsize}\selectfont}%
  \ifx\svgwidth\undefined%
    \setlength{\unitlength}{188.81835273bp}%
    \ifx\svgscale\undefined%
      \relax%
    \else%
      \setlength{\unitlength}{\unitlength * \real{\svgscale}}%
    \fi%
  \else%
    \setlength{\unitlength}{\svgwidth}%
  \fi%
  \global\let\svgwidth\undefined%
  \global\let\svgscale\undefined%
  \makeatother%
  \begin{picture}(1,1.05322096)%
    \lineheight{1}%
    \setlength\tabcolsep{0pt}%
    \put(0,0){\includegraphics[width=\unitlength,page=1]{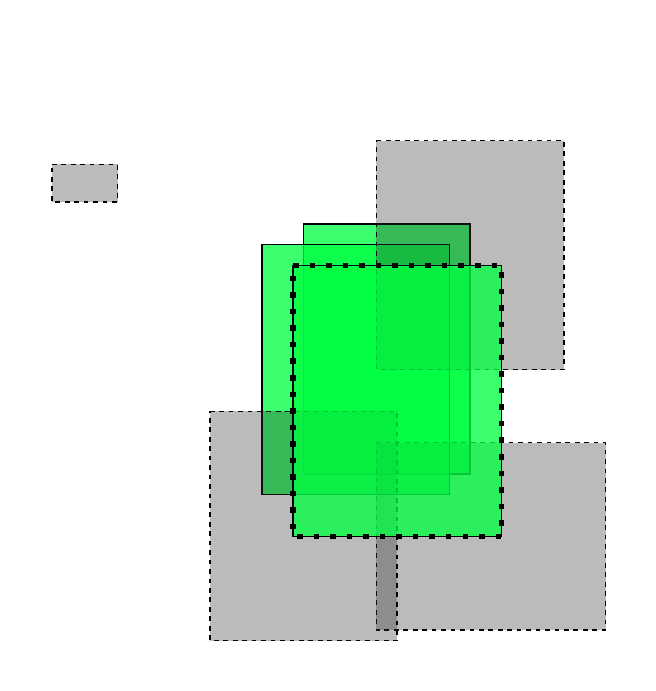}}%
    \put(0.20676375,0.75978563){\color[rgb]{0,0,0}\makebox(0,0)[lt]{\lineheight{1.25}\smash{\begin{tabular}[t]{l}$f(\bm{X}) \setminus \mathrm{Prop}(\widehat{b}^i)$\end{tabular}}}}%
    \put(0,0){\includegraphics[width=\unitlength,page=2]{../figs/proposal_illustration.pdf}}%
    \put(0.20676375,0.84577549){\color[rgb]{0,0,0}\makebox(0,0)[lt]{\lineheight{1.25}\smash{\begin{tabular}[t]{l}$\mathrm{Prop}(\widehat{b}^i)$\end{tabular}}}}%
    \put(0,0){\includegraphics[width=\unitlength,page=3]{../figs/proposal_illustration.pdf}}%
    \put(0.20676374,0.92712737){\color[rgb]{0,0,0}\makebox(0,0)[lt]{\lineheight{1.25}\smash{\begin{tabular}[t]{l}$\widehat{b}^i$\end{tabular}}}}%
  \end{picture}%
\endgroup%

    }
    \caption{Left: Illustration of the $P^i$ and $\varPhi^i$ features counting Lidar points falling into a given predicted box. From the points $\bm{X} \cap \widehat{b}^i$, reflection statistics are generated. Right: Schematic illustration of the proposal set $\mathrm{Prop}(\widehat{b}^i)$ for a given predicted box $\widehat{b}^i$ (here, in two dimensions for simplicity).
    From the proposal boxes, further pre-NMS statistics are derived.}\label{fig:md3d-feature-illustration}
\end{figure}

\paragraph{LMD Features.}
From this information we generate geometrical and statistical features for each $\widehat{b}^i \in \NMS[f(\bm{X})]$ for the purpose of UQ.
In addition to the bounding box features 
\begin{equation}
	\widehat{\phi}^i := \{\widehat{x}^i, \widehat{y}^i, \widehat{z}^i, \widehat{\ell}^i, \widehat{w}^i, \widehat{h}^i, \widehat{\theta}^i, \widehat{s}^i, \widehat{\kappa}^i\}
\end{equation}
of $\widehat{b}^i$ we compute the geometric features \emph{volume} $V^i = \widehat{\ell}^i \widehat{w}^i \widehat{h}^i$, \emph{surface area} $A^i = 2 (\widehat{\ell}^i \widehat{w}^i + \widehat{\ell}^i \widehat{h}^i + \widehat{w}^i \widehat{h}^i)$, \emph{relative size} $F^i = V^i / A^i$, \emph{number of Lidar points} $P^i = |\bm{X} \cap \widehat{b}^i|$ within $\widehat{b}^i$ and \emph{fraction of Lidar points} $\varPhi^i = P^i/ |\bm{X}|$ in $\widehat{b}^i$, see \cref{fig:md3d-feature-illustration} on the left for an illustration.
Moreover, each Lidar point that falls into $\widehat{b}^i$ (i.e., in $\bm{X} \cap \widehat{b}^i$) has a reflectance value $r$.
We add the maximal ($\rho_{\max}^i$), mean ($\rho_\mathrm{mean}^i$) and standard deviation ($\rho_\mathrm{std}^i$) over all reflectance values of points in $\widehat{b}^i$.
Lastly, for each $\widehat{b}^i$, we take the pre-NMS statistics into consideration which involves all proposal boxes in $f(\bm{X})$ that are NMS-filtered by $\widehat{b}^i$, i.e., the pre-image
\begin{equation}
    \mathrm{Prop}(\widehat{b}^i) := \NMS^{-1} [ \{ \widehat{b}^i\} ].
\end{equation}
These are characterized by having a significant three-dimensional $\iouthreed$ with $\widehat{b}^i$, see \cref{fig:md3d-feature-illustration} on the right.
\begin{figure}
    \begin{center}
        \small
        \resizebox{\textwidth}{!}{
            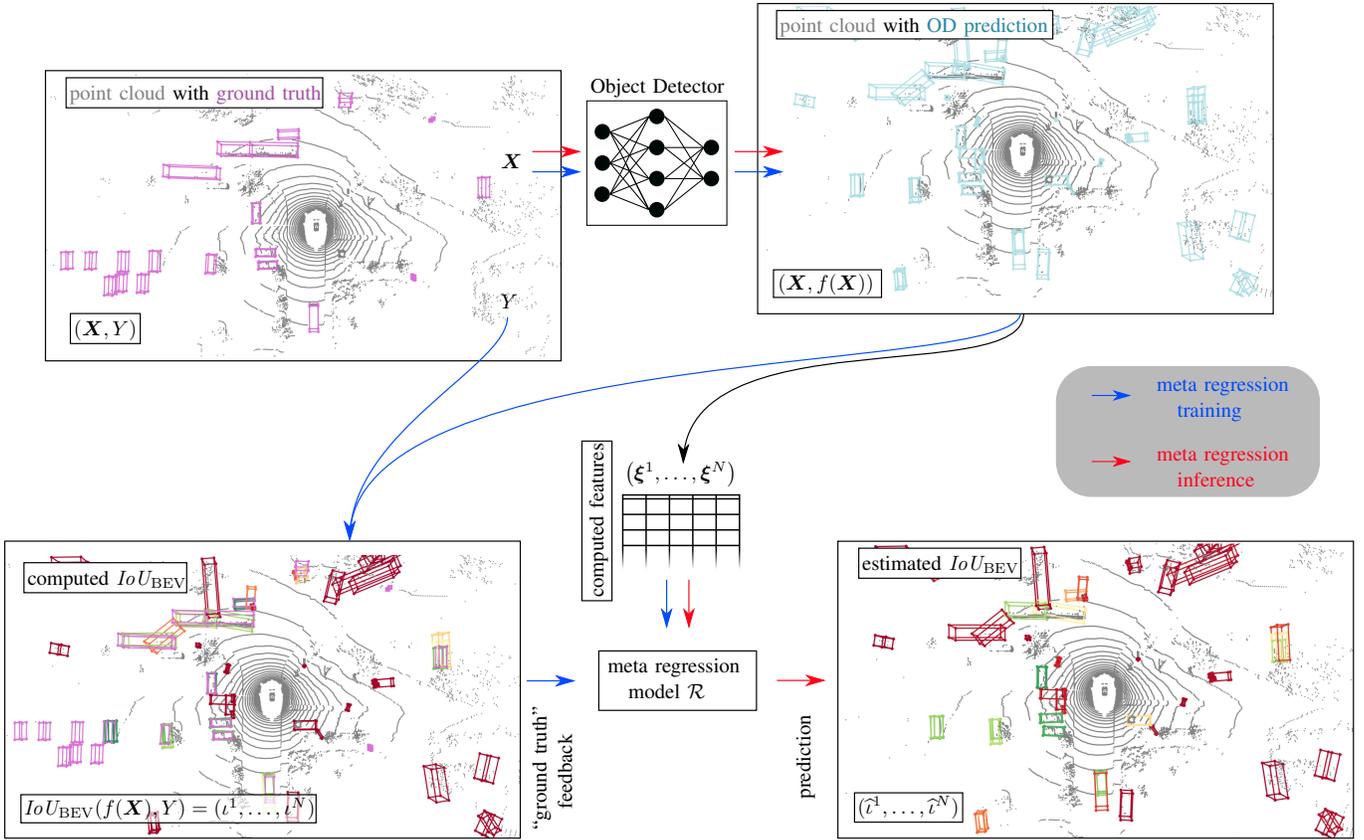
        }
    \end{center}
    \caption{Schematic illustration of the LMD meta regression pipeline.
    Training of the model is based on the output $f(\bm{X})$ of a fixed (frozen) object detector and the bounding box ground truth $Y$.
    Meta classification follows the same scheme with binary training targets $\tau^i = 1_{\{\iota^i > 0.5\}}$.
    }\label{fig: metareg-illustration}
\end{figure}
The number of proposal boxes $N^i := |\mathrm{Prop}(\widehat{b}^i)|$ is an important statistics since regions with more proposals are more likely to contain a true prediction.
We further derive minimum, maximum, mean and standard deviation statistics over proposal boxes $\widehat{b} \in \mathrm{Prop}(\widehat{b}^i)$ for all
\begin{equation}
    m^i \in \widehat{\phi}^i \cup \{V^i, A^i, F^i, P^i, \varPhi^i, \rho_{\max}^i, \rho_\mathrm{mean}^i, \rho_{\mathrm{std}}^i\},
\end{equation}
as well, as the $\iouthreed$ and bird-eye intersection over union $\ioubev$ values between $\widehat{b}^i$ and all proposals $\mathrm{Prop}(\widehat{b}^i)$.
Overall, this amounts to a vector $\bm{\xi}^i(\bm{X})$ of length $n = 90$ consisting of co-variables (features) on which post-processing models are fit in order to predict the $\iou_{\mathrm{BEV}}$ between $\widehat{b}^i$ and the ground truth or classify samples as true (TP) or false positives (FP).
We call a box a TP if $\iou_{\mathrm{BEV}} \geq 0.5$, otherwise we declare it FP.

\paragraph{Post-Processing.}
On an annotated hold-out dataset $\mathcal{D}_\mathrm{val}$ (consisting of point cloud-annotation tuples $(\bm{X}, Y)$), we compute a structured dataset $\mathsf{X} = (\bm{\xi}^1, \ldots, \bm{\xi}^{N_\mathrm{val}}) \in \R^{n \times N_\mathrm{val}}$ consisting of feature vectors for each of the $N_\mathrm{val}$ predicted boxes over all of $\mathcal{D}_\mathrm{val}$.
The illustration of our method in \cref{fig: metareg-illustration} shows this scheme for one particular Lidar frame $(\bm{X}, Y)$ and the respective prediction on it.
Further, we compute $\iota^i := \ioubev (\widehat{b}^i(\bm{X}), Y)$ between prediction and ground truth form $\mathcal{D}_\mathrm{val}$ as target variables $\mathsf{Y} = (\iota^1, \ldots, \iota^{N_\mathrm{val}}) \in \R^{N_\mathrm{val}}$.
We then fit a light-weight \emph{(meta-) regression model} \(\mathcal{R}: \bm{\xi}^i \mapsto \mathsf{Y}_i\) on $(\mathsf{X}, \mathsf{Y})$ which acts as post-processing module of the detector in order to produce $\ioubev$-estimates $\widehat{\iota}^i := \mathcal{R}(\bm{\xi}^i )$ for each detection $\widehat{b}^i$.
Similarly, we fit a binary \emph{(meta-) classification model} $\mathcal{C}$ obtaining the binary targets $1_{\{\mathsf{Y} > 0.5\}}$ which allows us to generate alternative confidence estimates $\widehat{\tau}^i := \mathcal{C}(\bm{\xi}^i) \in [0, 1]$ for each prediction $\widehat{b}^i$ in post-processing.
Note that $\mathcal{C}$ is a potentially non-linear and non-monotonous function of the features $\bm{\xi}^i$ and, therefore, can change the obtained confidence ranking per frame and influence detection performance as opposed to simple re-calibration methods~\cite{naeini_mahdi_pakdaman_obtaining_2015,guo_calibration_nodate}.

Meta classification empirically turns out to produce confidence estimates which are both, sharper (in the sense of separating TPs from FPs) and better calibrated that those produced natively by the detector, i.e., the objectness score.
However, when regarding the cases of disagreement between the computed $\ioubev$ and \(\mathcal{C}\), we frequently find that \(\mathcal{C}\) is to be trusted more than the computed $\ioubev$ due to missing annotations.
We use this observation in order to generate proposals (in descending estimation $\widehat{\tau}^i$) based on the object detector in comparison with the given ground truth (FP according to the ground truth, i.e., $\iota^i < 0.5$) that serve as suggestions of annotation errors.

\section{Numerical Results}
In this section we study meta classification and meta regression performance for two benchmark datasets as well as a proprietary dataset by \aptiv.
The meta classification results are presented in terms of accuracy and area under receiver operator characteristics curve ($\auroc$~\cite{davis_relationship_2006}) and the meta regression results are presented in terms of $R^2$.
We compare our uncertainty quantification method LidarMetaDetect (LMD) with two baseline methods (score, box features).
Moreover, we detect annotation errors on both benchmark datasets using LMD.

\paragraph{Implementation Details.}
We implemented our method in the open source MMDetection3D toolbox~\cite{mmdetection3d_contributors_openmmlabs_2020}.
For our experiments, we consider the PointPillars~\cite{lang_pointpillars_2019} and CenterPoint~\cite{yin_center-based_2021} architectures.
The mean average precision ($\map @ \iou_{\!0.5}$) for KITTI based on $\ioubev$ is $69.0$ for CenterPoint and $68.8$ for PointPillars. 
On KITTI, the $\map @ \iou_{\!0.5}$ based on $\iouthreed$ is $64.2$ for CenterPoint and $68.8$ for PointPillars and for \aptiv, the $\map @ \iou_{\!0.5}$ based on $\iouthreed$ is $39.5$ for CenterPoint and $43.7$ for PointPillars.
NuScenes performance is given as a weighted sum of $\map$ as well as the nuScenes detection score (NDS).
For CenterPoint, the $\map$ is $57.4$ and the NDS is $65.2$ and for PointPillars the $\map$ is $40.0$ and the NDS is $53.3$.
For KITTI and \aptiv, the models were trained individually while available public model weights from MMDetection3D are used for nuScenes.
The performance results obtained have all been evaluated on respective test datasets.
For KITTI, the images and associated point clouds are split scene-wise, such that the training set consists of $3,\!712$, the validation set of $1,\!997$, and the test set of $1,\!772$ frames.
For nuScenes, the validation set is split scene-wise into $3,\!083$ validation and $2,\!936$ test frames.
The \aptiv{} dataset consists of $50$ sequences, split into $27, 14, 9$ sequences with about $145$K, $75$K, $65$K cuboid annotations for training, validation and testing, respectively. Every sequence is about two minutes long while every fifth point cloud is annotated. The covered locations are countryside, highway and urban from and around (anonymous). The dataset includes four classes: $1.$ smaller vehicles likes cars and vans, $2.$ larger vehicles like busses and trucks, $3.$ pedestrians and $4.$ motorbikes and bicycles.

\paragraph{Correlation of Box-wise Features with the $\ioubev$.}
\begin{figure}
    \centering
    \begin{minipage}{0.53\textwidth}
        \resizebox{0.8\textwidth}{!}{
        \begin{tikzpicture}
            \begin{axis}[ybar,
                ylabel=Pearson correlation with \(\ioubev\),
                xtick=data,
                x tick label style=
                    {rotate=45,anchor=east,font=\small},
                symbolic x coords={$\max\{s^{i}\}$, $\mathrm{std}\{s^{i}\}$, $s^i$, $\mathrm{mean}\{s^{i}\}$, $N^{i}$, $\max\{\iouthreed^i\}$, $\mathrm{mean}\{\iouthreed^{i}\}$, $\mathrm{std}\{\rho^{i}_\mathrm{max}\}$, $\mathrm{std}\{\iouthreed^{i}\}$, $\max\{\rho^{i}_\mathrm{max}\}$, $\max\{P^{i}\}$, $\mathrm{mean}\{P^{i}\}$}
                ]
            \addplot coordinates
                {($\max\{s^{i}\}$,0.7516) ($\mathrm{std}\{s^{i}\}$,0.6991) ($s^i$, 0.6755) ($\mathrm{mean}\{s^{i}\}$, 0.5847) ($N^{i}$, 0.3007) ($\max\{\iouthreed^i\}$, 0.2900) ($\mathrm{mean}\{\iouthreed^{i}\}$, 0.2707) ($\mathrm{std}\{\rho^{i}_\mathrm{max}\}$, 0.2652) ($\mathrm{std}\{\iouthreed^{i}\}$, 0.2560) ($\max\{\rho^{i}_\mathrm{max}\}$, 0.2556) ($\max\{P^{i}\}$, 0.2519) ($\mathrm{mean}\{P^{i}\}$, 0.2500)
                };
            \end{axis}
        \end{tikzpicture}
        }
        \caption{Strongest correlation coefficients for constructed box-wise features and $\ioubev$ for the CenterPoint architecture on the nuScenes test dataset and a score threshold $\tau$ = 0.1.}
        \label{fig:pearson-correlation-with-ioubev}
    \end{minipage}
    \hfill
    \begin{minipage}{0.43\textwidth}
        \resizebox{0.8\linewidth}{!}{
        \begin{tikzpicture}
            \begin{axis}[
                ylabel=$\auroc$,
                height=4cm,
                width=6.5cm
                ]
            \addplot[color=blue,mark=*] coordinates {
                (1, 0.9253) (2, 0.9417) (3, 0.9591) (4, 0.9603) (5, 0.9609) (6, 0.9611) (7, 0.9614) (8, 0.9617) (9, 0.9619) (10, 0.9620)
            };
            \addplot[teal,sharp plot,dashed,update limits=false] 
        coordinates {(-2, 0.9628) (12,0.9628)};
            \end{axis}
        \end{tikzpicture}
        }
        \resizebox{0.8\linewidth}{!}{
        \begin{tikzpicture}
            \begin{axis}[
                xlabel=Number of features,
                ylabel=$R^2$,
                height=4cm,
                width=6.5cm
                ]
            \addplot[color=blue,mark=*] coordinates {
                (1, 0.5892) (2, 0.6591) (3, 0.7079) (4, 0.7158) (5, 0.7204) (6, 0.7243) (7, 0.7264) (8, 0.7280) (9, 0.7286) (10, 0.7289)
            };
            \addplot[teal,sharp plot,dashed,update limits=false] 
        coordinates {(-2, 0.7296) (12, 0.7296)};
            \end{axis}
        \end{tikzpicture}
        }
        \caption{Feature selection using a greedy heuristic for CenterPoint, nuScenes and score threshold $0.1$. The top figure contains meta classification $\auroc$ and the bottom one contains meta regression $R^2$. The dashed line shows the the performance when incorporating all features (LMD).}\label{fig:feature-selection-lmd}
    \end{minipage}
\end{figure}
\Cref{fig:pearson-correlation-with-ioubev} shows the Pearson correlation coefficients of the constructed box-wise dispersion measures with the $\ioubev$ of prediction and ground truth for CenterPoint on the nuScenes test dataset. 
The score features have strong correlations ($> 0.5$) with the $\ioubev$.
Note that, although the four score-related features show the highest individual correlation, these features may be partially redundant.
The number of candidate boxes $N^{i}$ is also reasonably correlated with the $\ioubev$ ($0.3007$), whereas the remaining features only show a minor correlation ($< 0.3$).
However, they may still contribute to higher combined explanatory information in meta classification.
The strongest correlation for the other network-dataset combinations are shown in the appendix.

\paragraph{Comparison of Different Meta Classifiers and Regressors.}
\begin{table*}[t]
    \centering
    \caption{Comparison of meta classification accuracy and $\auroc$ as well as meta regression $R^2$ values for the score baseline, bounding box features and LMD for CenterPoint and nuScenes test dataset with score threshold $0.1$; higher values are better. Bold numbers indicate the highest performance and underlined numbers represent the second highest (row-wise). 
    Models used are Logistic Regression (LogReg), Ridge Regression (RR), Random Forest (RF), Gradient Boosting (GB) and a Multi Layer Perceptron (MLP).
    }
    \resizebox{\textwidth}{!}{
    \begin{tabular}{c | c c c c | c c c c | c c c c}\toprule
         & \multicolumn{8}{|c}{Meta Classification} & \multicolumn{4}{|c}{Meta Regression}  \\
        \toprule
        & \multicolumn{4}{|c}{Accuracies} & \multicolumn{4}{|c}{$\auroc$s} & \multicolumn{4}{|c}{$R^2$} \\
         \midrule
         Method & LogReg & RF & GB & MLP & LogReg & RF & GB & MLP & RR & RF & GB & MLP \\
          \midrule
          Score & \textbf{0.8777} & 0.8524 & 0.8772 & \underline{0.8773} & \textbf{0.8644} & 0.8617 & 0.8623 & \underline{0.8640} & 0.4641 & 0.4675 & \underline{0.4733} & \textbf{0.4751} \\ 
          Box Features & 0.8877 & \underline{0.9049} & \textbf{0.9203} & 0.8975 & 0.9056 & \underline{0.9454} & \textbf{0.9529} & 0.9293 & 0.5292 & \underline{0.6681} & \textbf{0.6792} & 0.6249 \\ 
          LMD & 0.9118 & 0.9166 & \textbf{0.9297} & \underline{0.9200} & 0.9450 & \underline{0.9581} & \textbf{0.9628} & 0.9530 & 0.6451 & \underline{0.7242} & \textbf{0.7296} & 0.7122 \\  
          \bottomrule
        \end{tabular}
        }
    \label{tab: meta model comparison}
\end{table*}
Different models can serve as post-processing modules for meta classification ($\mathcal{C}$) and meta regression ($\mathcal{R}$, see \cref{sec:method}).
For meta classification, the meta models under consideration are logistic regression (LogReg), random forest (RF), gradient boosting (GB) and a multilayer perceptron (MLP) with two hidden layers.
For meta regression, analogous regression models are used, only the logistic regression is replaced with a ridge regression (RR). 

The respective meta models are trained on the box-wise features $\bm{\xi}^i$ of the validation sets $\mathcal{D}_\mathrm{val}$ and evaluated on the features of the test sets which are disjoint from $\mathcal{D}_\mathrm{val}$.
LMD uses all available features to train the meta models, whereas in the score baseline only the score of the prediction $\widehat{s}^i$ is used to fit the meta model. 
For the bounding box features baseline, the box features of the prediction $\widehat{\phi}^i$ are used, in which the score $\widehat{s}^i$ is also included.
\Cref{tab: meta model comparison} presents meta classification accuracy and $\auroc$ as well as meta regression $R^2$ for the CenterPoint architecture on the nuScenes dataset.
For the score baseline, all meta models perform similarly well.
For the meta classification accuracy there are differences of up to $2.53$ percent points (pp), for the $\auroc$ of at most $0.27$ pp and for meta regression $R^2$ of up to $1.10$ pp.
For the box features the maximum differences increase to $3.26$ pp in terms of accuracy, to $4.73$ pp for $\auroc$ and to $15.00$ pp for $R^2$.
In particular, for the box features and LMD, the non-linear models (RF, GB, MLP) outperform the linear model in both learning tasks.
LMD outperforms the baselines box features/score by $0.94/5.20$ pp in terms of accuracy, by $0.99/9.84$ pp in terms of $\auroc$ and by $5.04/25.45$ pp in terms of $R^2$.
If overfitting of the meta model is made unlikely by choosing appropriate hyperparameters, the performance of the meta model typically benefits from adding more features, since the available information and number of parameters for fitting are increased.
Overall, GB outperforms all other meta models, especially when multiple features are used to train and evaluate the respective learning task.
Therefore, only results based on GB are shown in the following experiments. 
An overview of the different meta models for all network-dataset combinations are shown in the appendix.

\paragraph{Comparison for Different Datasets and Networks.}
\begin{table*}[t]
    \centering
    \caption{Comparison of meta classification accuracy and $\auroc$ as well as meta regression $R^2$ for the score baseline, bounding box features and LMD for all available network-dataset combinations with $\ioubev$ threshold $0.5$, score threshold $0.1$ and GB as meta model.
    Higher values are better. Bold numbers indicate the highest performance and underlined numbers represent the second highest (row-wise).}
    \resizebox{0.85\textwidth}{!}{
    \begin{tabular}{c c | c c c | c c c | c c c }\toprule
         &  & \multicolumn{6}{c}{Meta Classification} & \multicolumn{3}{|c}{Meta Regression} \\
        \toprule
         & & \multicolumn{3}{|c}{Accuracies} & \multicolumn{3}{|c}{$\auroc$s} & \multicolumn{3}{|c}{$R^2$} \\
         \midrule
         Dataset & Network & Score & Box & LMD & Score & Box & LMD & Score & Box & LMD  \\
         \midrule
        \multirow{2}{*}{KITTI} & PointPillars & 0.8921 & \underline{0.8931} & \textbf{0.9004} & 0.9530 & \underline{0.9537} & \textbf{0.9592} & 0.7108 & \underline{0.7131} & \textbf{0.7287} \\ 
         & CenterPoint & 0.8688 & \underline{0.8691} & \textbf{0.8806} & 0.9274 & \underline{0.9343} & \textbf{0.9466} & 0.6235 & \underline{0.6472} & \textbf{0.6840} \\ 
         \midrule
         \multirow{2}{*}{nuScenes} & PointPillars & 0.8398 & \underline{0.8708} & \textbf{0.8915} & 0.8129 & \underline{0.9002} & \textbf{0.9280} & 0.4055 & \underline{0.5593} & \textbf{0.6413} \\ 
         & CenterPoint & 0.8772 & \underline{0.9203} & \textbf{0.9297} & 0.8623 & \underline{0.9529} & \textbf{0.9628} & 0.4732 & \underline{0.6792} & \textbf{0.7296} \\
         \midrule
         \multirow{2}{*}{\aptiv} & PointPillars & 0.7939 & \underline{0.8489} & \textbf{0.8615} & 0.8558 & \underline{0.9274} & \textbf{0.9396} & 0.5096 & \underline{0.6568} & \textbf{0.6924} \\ 
         & CenterPoint & 0.8265 & \underline{0.8440} & \textbf{0.8548} & 0.8914 & \underline{0.9134} & \textbf{0.9275} & 0.5456 & \underline{0.6286} & \textbf{0.6591} \\
        \bottomrule
    \end{tabular}
    }
    \label{tab: meta classification and regression}
\end{table*}
\Cref{tab: meta classification and regression} shows meta classification accuracy and $\auroc$ as well as meta regression $R^2$ for all network-dataset combinations based on GB models.
In all cases LMD outperforms both baselines and the bounding box features outperform the score baseline.
This is to be expected, since the score is contained in the box features and the box features are contained in the set of features of LMD.
The improvement from the score baseline to LMD ranges from $0.83$ to $6.76$ pp in terms of meta classification accuracy, from $0.62$ to $10.51$ pp in terms of $\auroc$ and from $1.79$ to $25.64$ pp in terms of meta regression $R^2$.
The improvement from the bounding box features to LMD ranges from $0.73$ to $2.07$ pp in terms of meta classification accuracy, from $0.55$ to $2.78$ pp in terms of $\auroc$ and from $1.56$ to $8.20$ pp in terms of meta regression $R^2$.
This illustrates that the addition of features, other than just the bounding box features of the prediction itself, has a significant impact on meta classification and meta regression performances and, therefore, separation of TP and FP predictions.

\begin{figure*}
    \centering
    \begin{minipage}{.475\textwidth}
        \includegraphics[width=\textwidth]{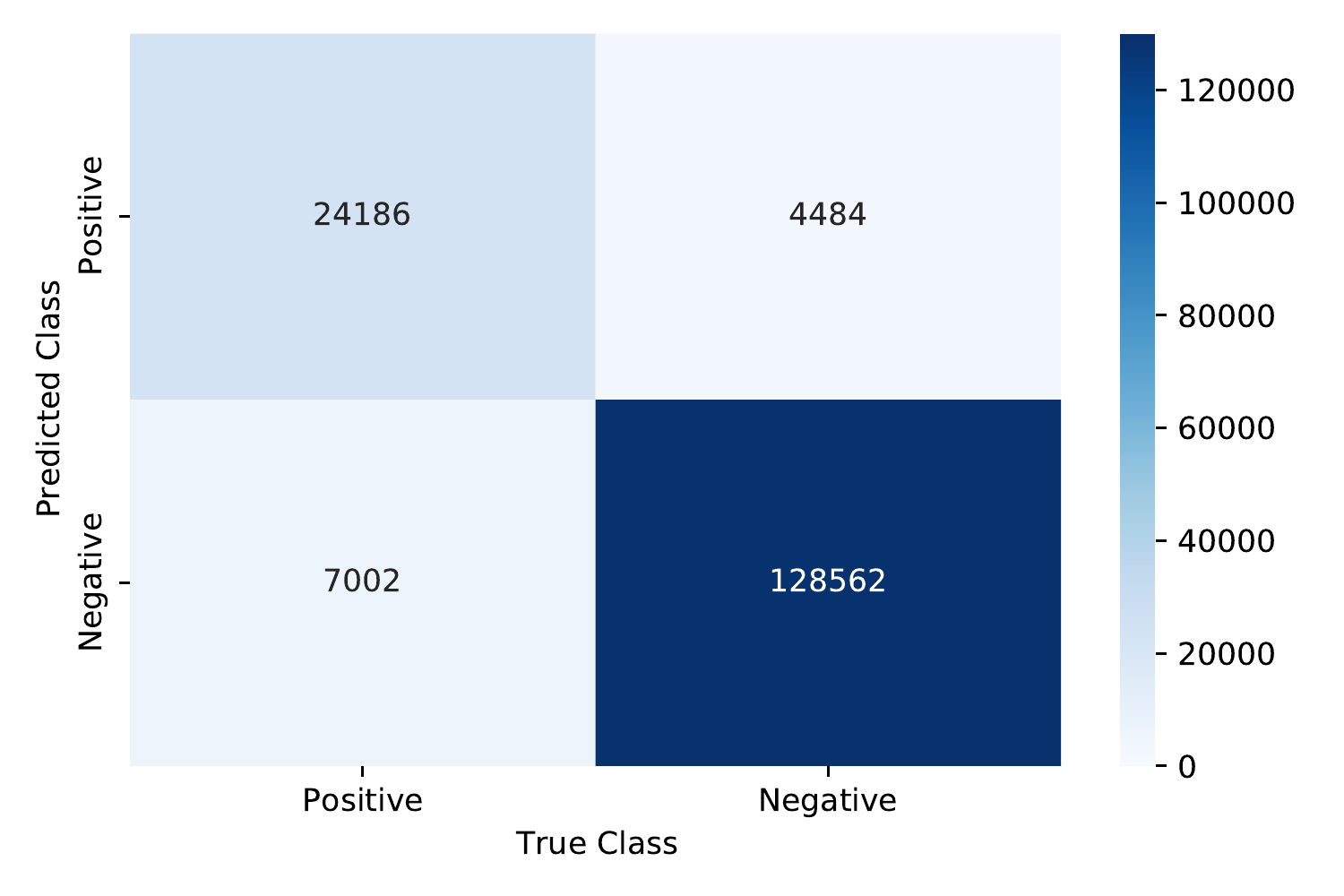}
        \caption{Confusion matrix of a GB classifier for LMD on CenterPoint, nuScenes and score threshold $0.1$.}
        \label{fig: confusion matrix}
    \end{minipage}
    \hfill
    \begin{minipage}{.475\textwidth}
        \includegraphics[width=\textwidth]{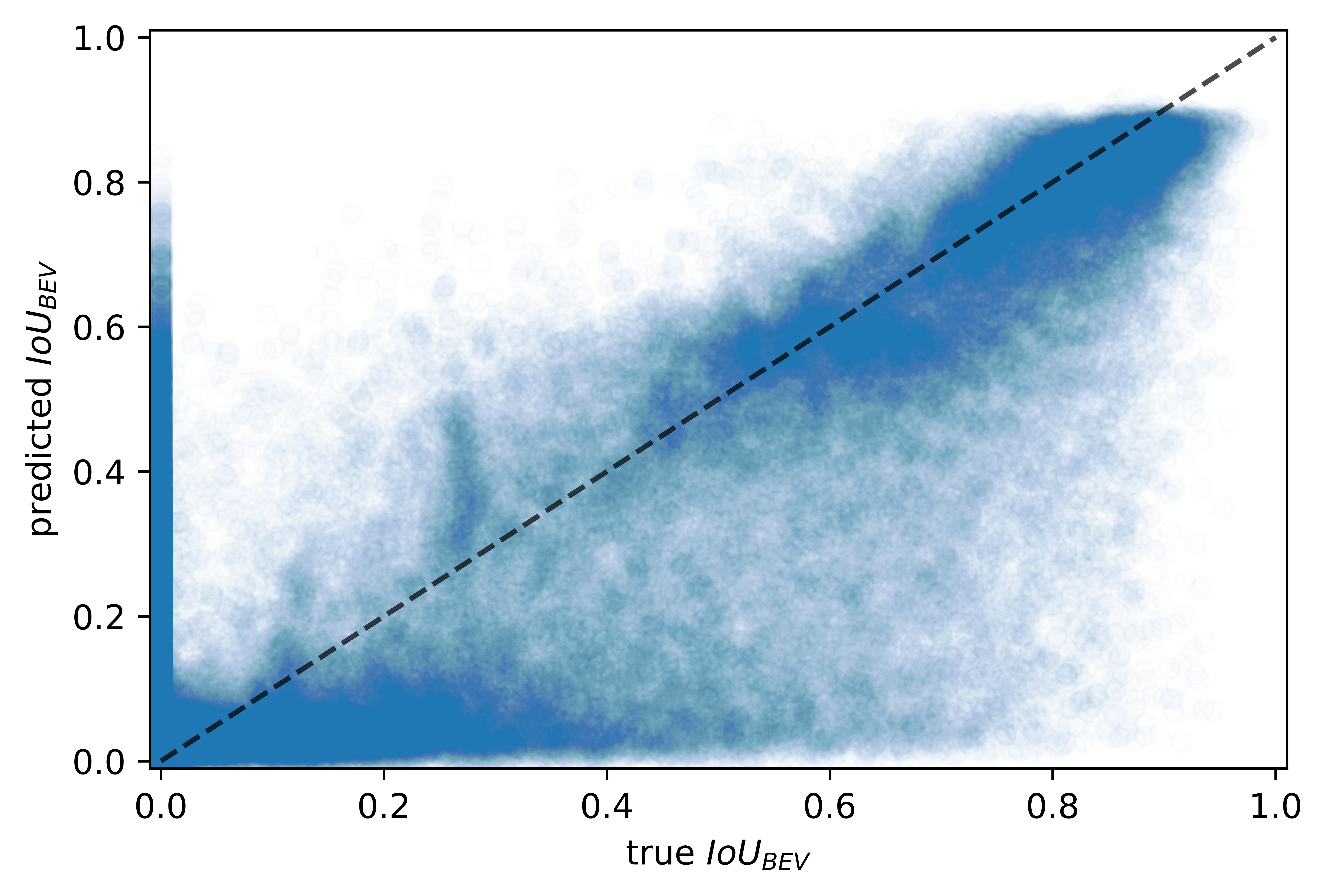}
        \caption{Box-wise scatter plot of true $\ioubev$ and predicted $\ioubev$ values for LMD on CenterPoint, nuScenes and score threshold $0.1$. The predictions are based on a GB regressor.}
        \label{fig: scatter plot}
    \end{minipage}
\end{figure*}

For CenterPoint and nuScenes, the confusion matrix \cref{fig: confusion matrix} shows that the GB classifier based on LMD identifies most TPs and true negatives.
Therefore, predictions that are in fact FPs are also predicted as FPs. 
Note, that here we regard ``meta'' true negatives conditional on the detectors prediction (each example is a detection TP or FP that is binarily classified).
The values on the off-diagonals indicate the errors of the meta classifier.
$7,\!002$ predictions are predicted as FPs even though they are TPs.
In contrast, $4,\!484$ predictions are predicted as TPs, even though they are actually FPs.
\Cref{fig: scatter plot} shows a scatter plot of the true $\ioubev$ of prediction and ground truth and the $\ioubev$ estimated by LMD meta regression based on a GB model, where each point represents one prediction.
Well-concentrated points around the identity (dashed line) indicate well-calibrated $\ioubev$-estimates and, therefore, object-wise quality estimates.

\paragraph{Feature Selection for Meta Classification and Meta Regression.}
Overall, LMD is based on $90$ features, which partly describe very similar properties.
In order to get a subset of features which contains as few redundancies as possible but is still powerful, we apply a greedy heuristic.
Starting with an empty set, a single feature that improves the meta prediction performance maximally is added iteratively. 
\Cref{fig:feature-selection-lmd} shows results in terms of $\auroc$ for meta classification and in terms of $R^2$ for meta regression for CenterPoint on nuScenes. 
The tests for the meta classification and the meta regression are independent of each other, i.e., the selected features of the two saturation plots do not have to match.
When using five selected features, the associated meta models perform already roughly as well as when using all features (LMD), i.e., $0.19$ pp worse in terms of meta classification $\auroc$ and $0.92$ pp worse in terms of meta regression $R^2$. 
With ten features used, the respective differences with the results obtained by LMD are $<0.1$ pp and thus negligible.
The tests for the greedy selection heuristic for all network-dataset-combinations are shown in the appendix.

\paragraph{Confidence Calibration.}
\begin{figure}
    \centering
    \includegraphics[width=0.55\columnwidth]{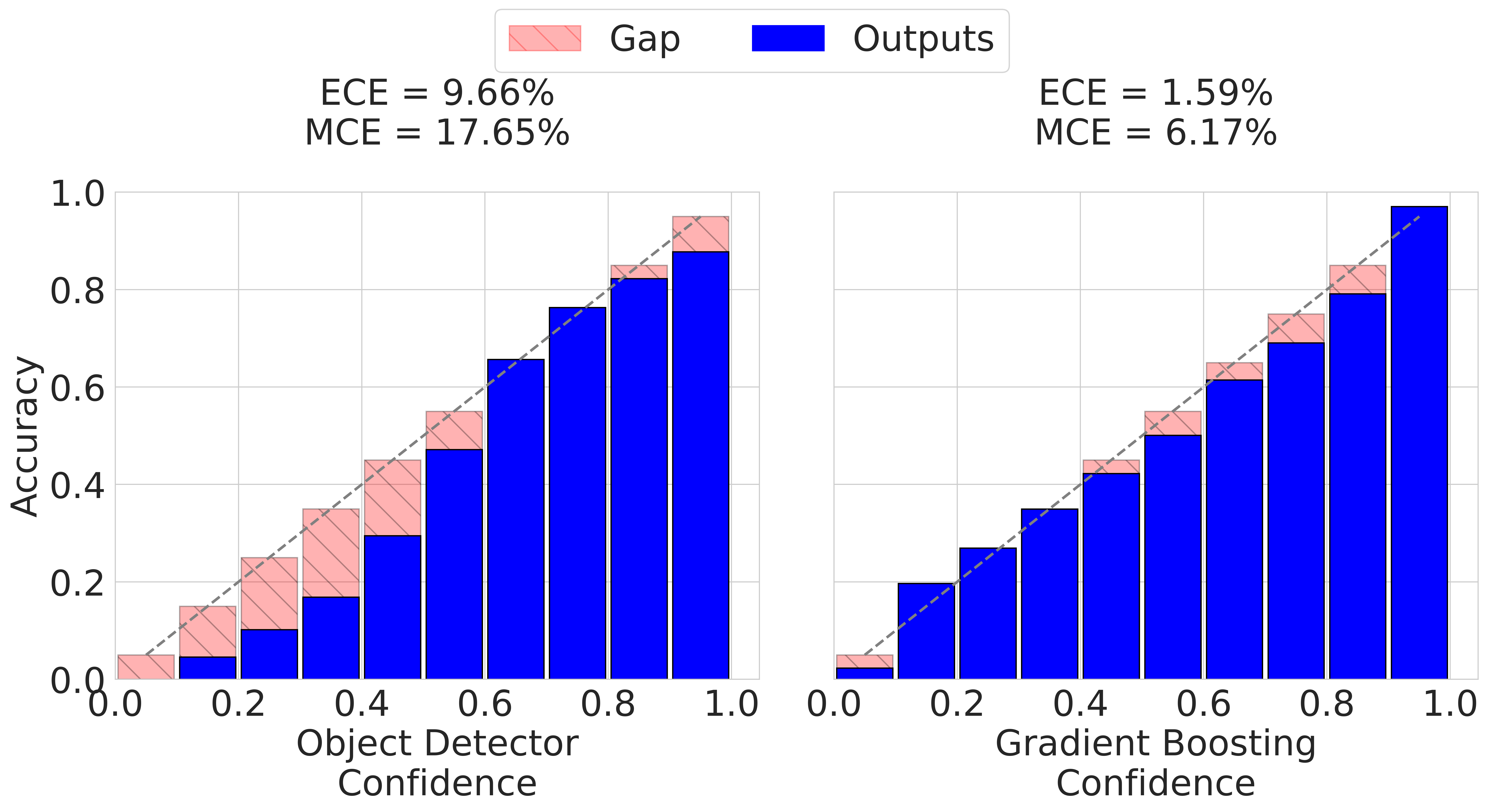}
    \caption{Reliability plots of the score (left) and GB classifier for LMD (right) with calibration errors (ECE, MCE) for CenterPoint, nuScenes test dataset, score threshold $0.1$ and $\ioubev$ threshold $0.5$.}
    \label{fig: calibration}
\end{figure}
The score and the meta classifier confidences are divided into 10 confidence bins to evaluate their calibration errors.
\Cref{fig: calibration} shows exemplary reliability plots for the object detector score and LMD based on a GB classifier with corresponding expected (ECE~\cite{naeini_mahdi_pakdaman_obtaining_2015}) and maximum calibration error (MCE~\cite{naeini_mahdi_pakdaman_obtaining_2015}).
The score is over-confident in the lower confidence ranges and well-calibrated in the upper confidence ranges, whereas the GB classifier for LMD is well-calibrated over all confidence ranges.
This observation is also reflected in the corresponding calibration errors, as the GB classifier for LMD outperforms the score by $8.07$ pp in terms of ECE and by $11.48$ pp in terms of MCE.
This indicates that LMD improves the statistical reliability of the confidence assignment.

        
        
        
        
        




\paragraph{Annotation Error Detection as an Application of Meta Classification.}
\begin{figure}
    \centering

    \includegraphics[width=0.5\textwidth,trim={10cm 1cm 2cm 4.35cm},clip]{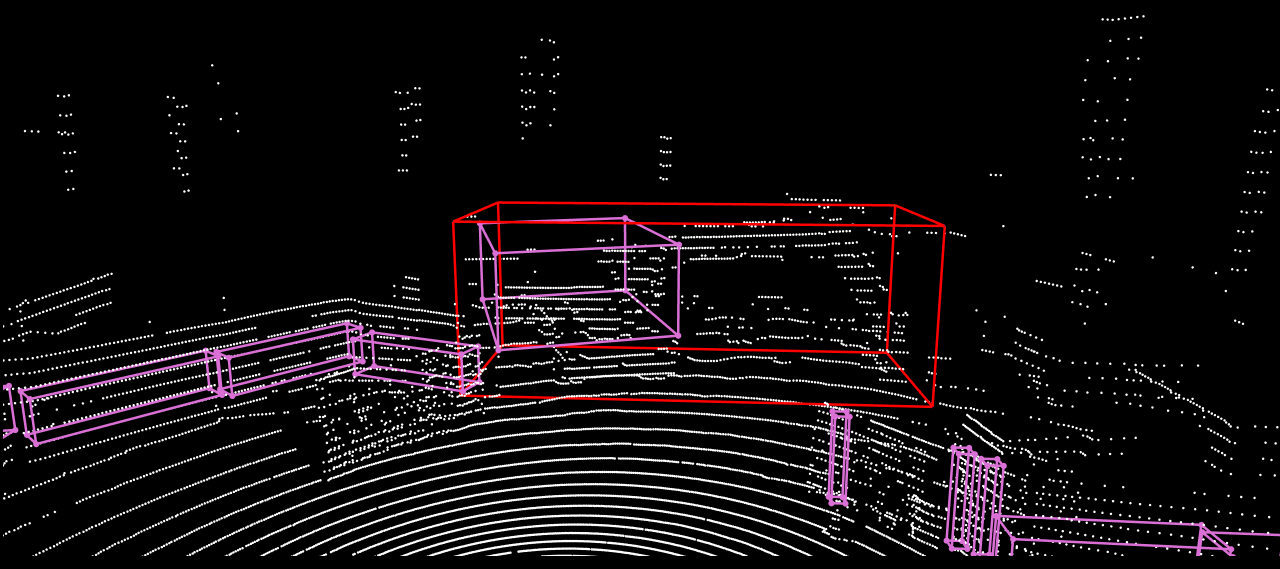}%
    \includegraphics[width=0.5\textwidth,trim={2cm 3cm 8cm 4.6cm},clip]{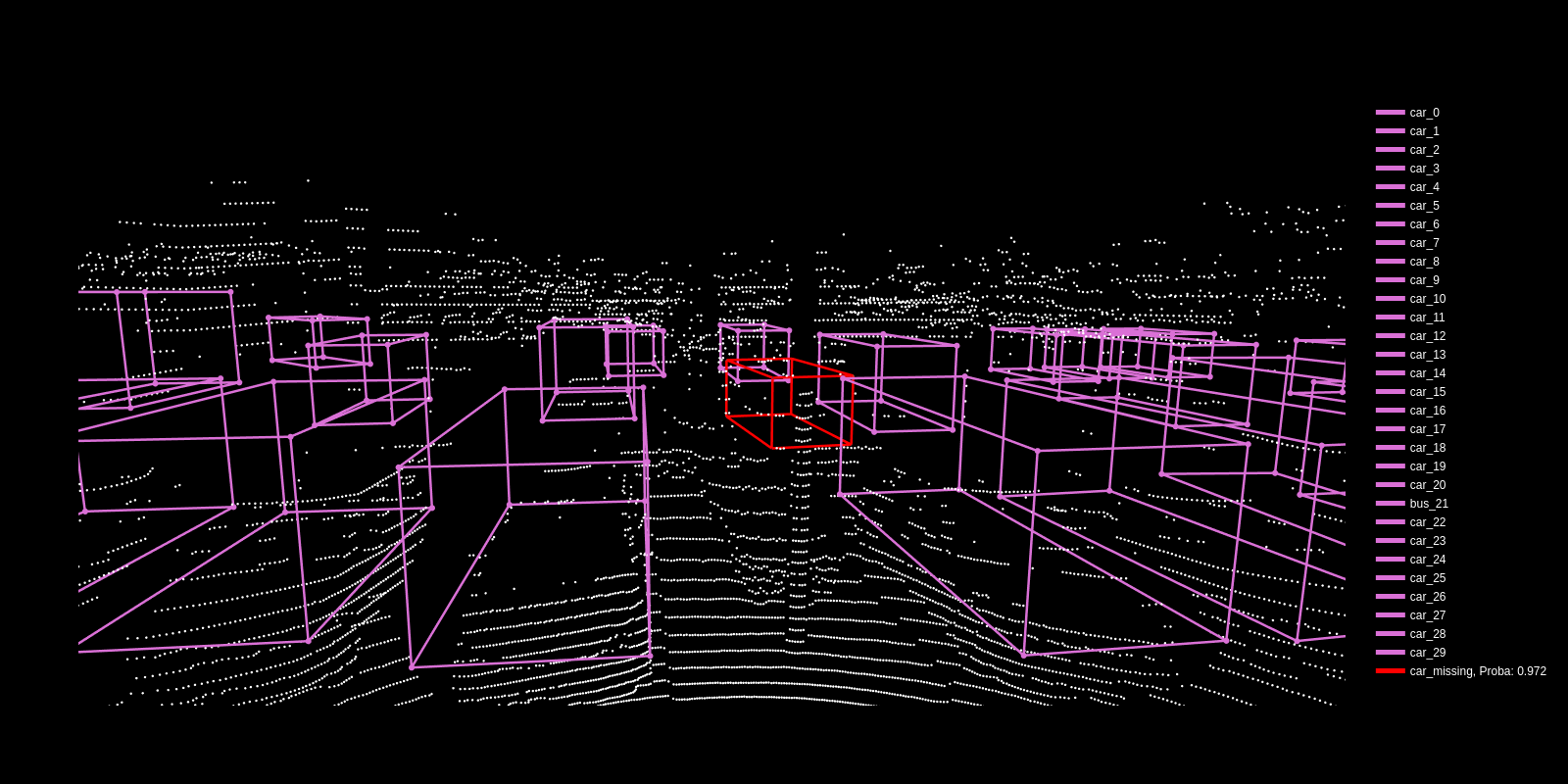}
    \includegraphics[width=0.5\textwidth,trim={0 1cm 0 2cm},clip]{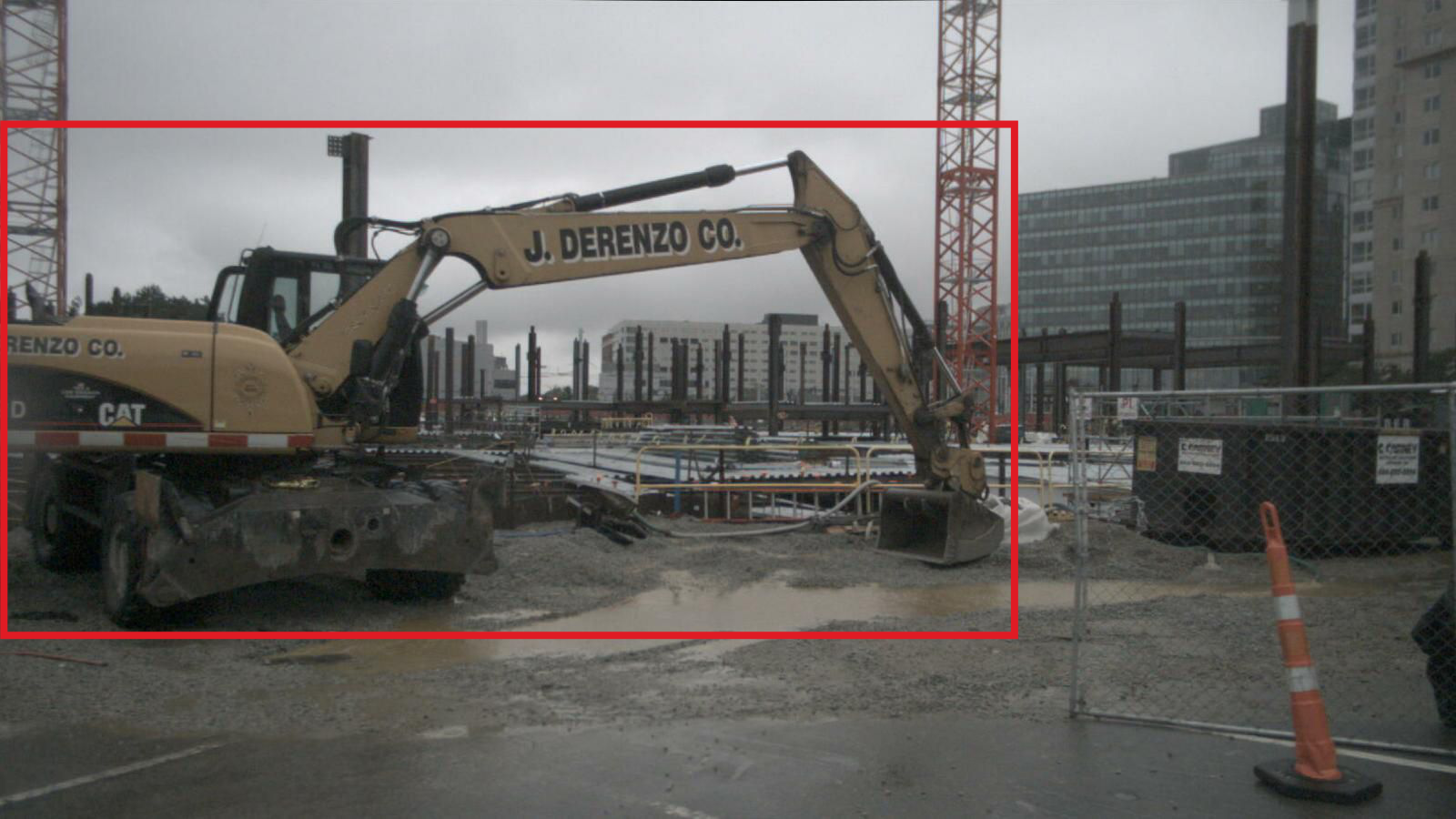}%
    \includegraphics[width=0.5\textwidth,trim={0 1cm 0 2cm},clip]{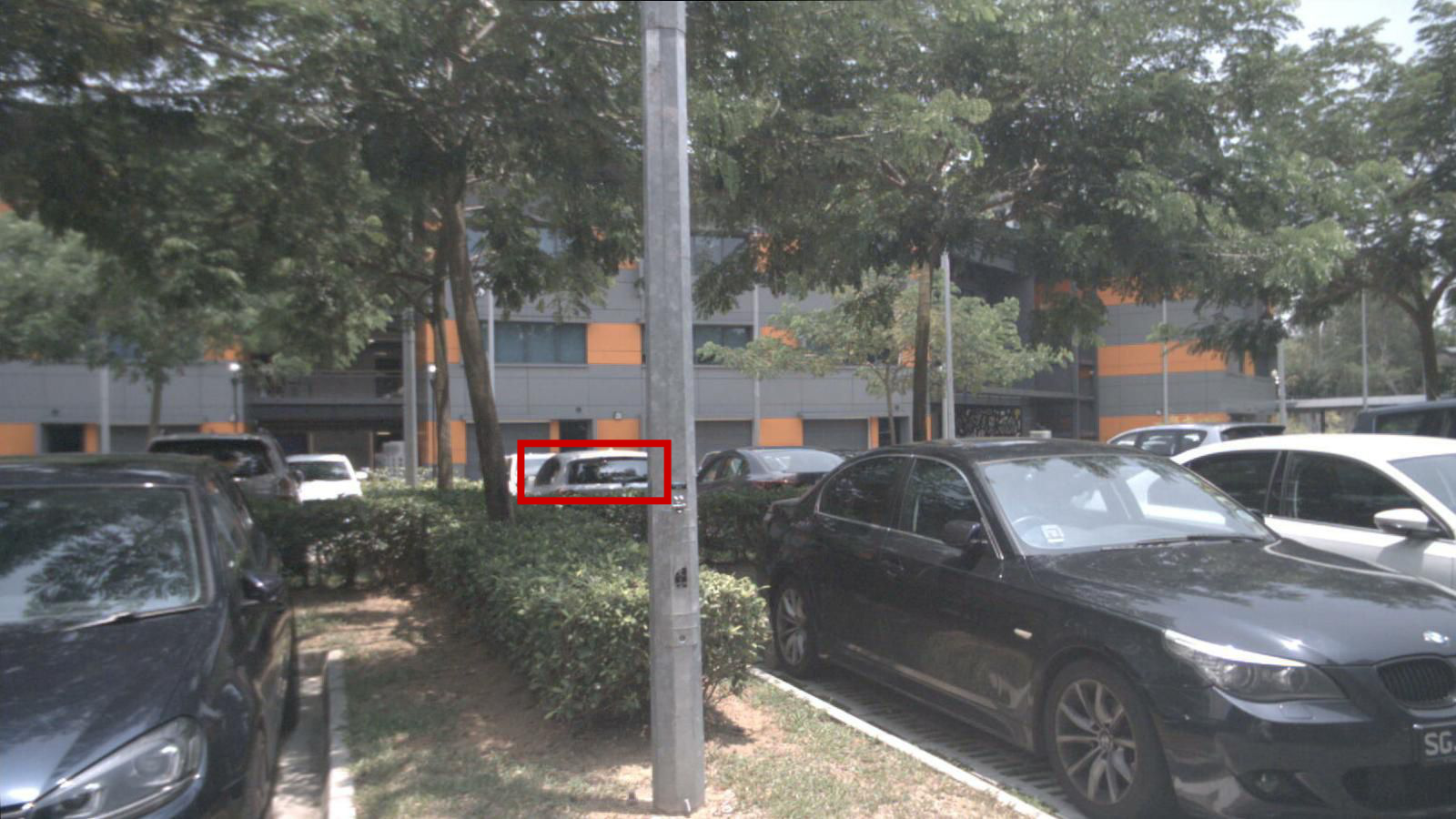}
    \includegraphics[width=0.5\textwidth,trim={9cm 5.5cm 6.84cm 4cm},clip]{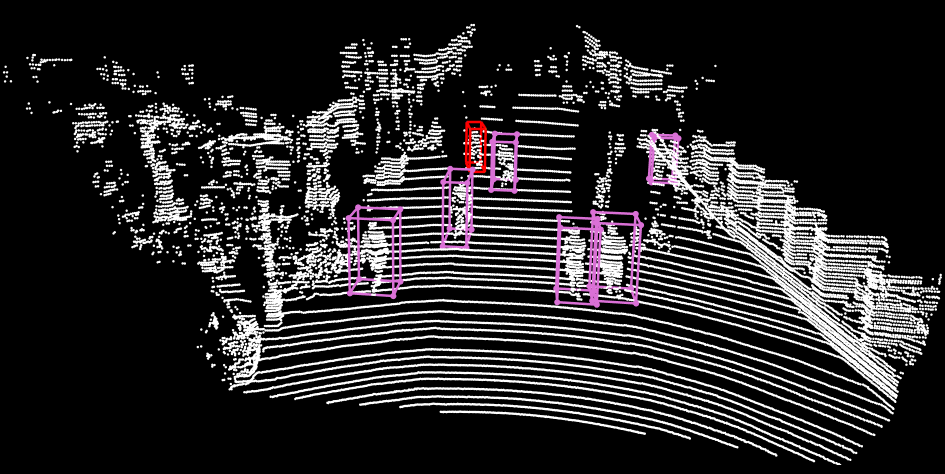}%
    \includegraphics[width=0.5\textwidth,trim={5cm 3.31cm 6cm 2cm},clip]{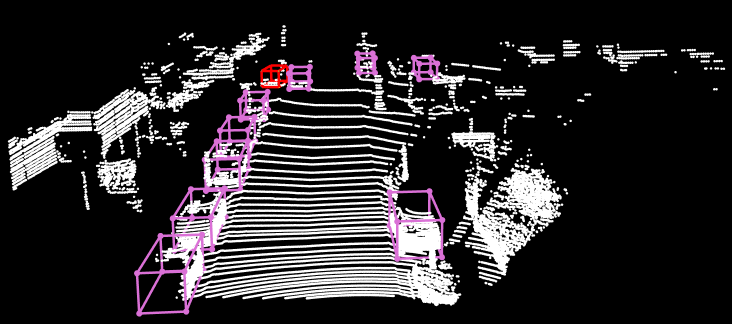}
    \includegraphics[width=0.5\textwidth,trim={7cm 2cm 11.5cm 4cm},clip]{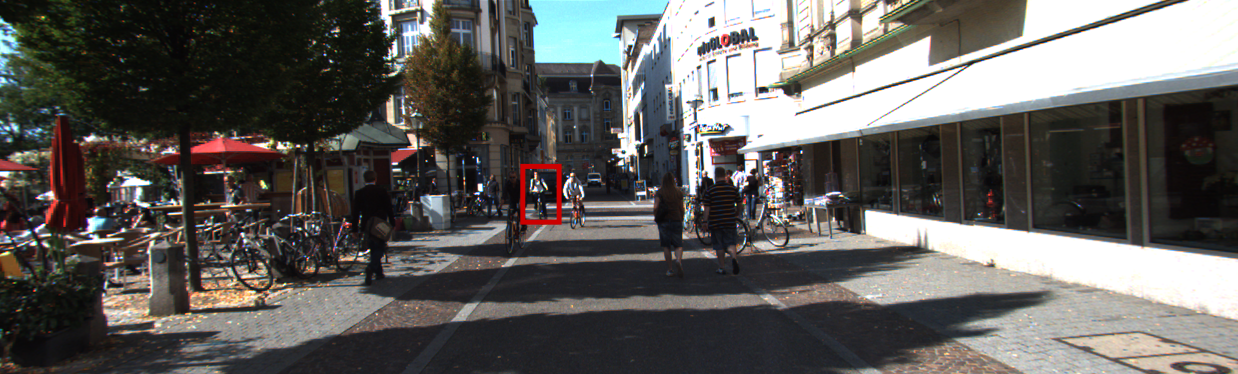}%
    \includegraphics[width=0.5\textwidth,trim={7.05cm 2.5cm 11.5cm 3.5cm},clip]{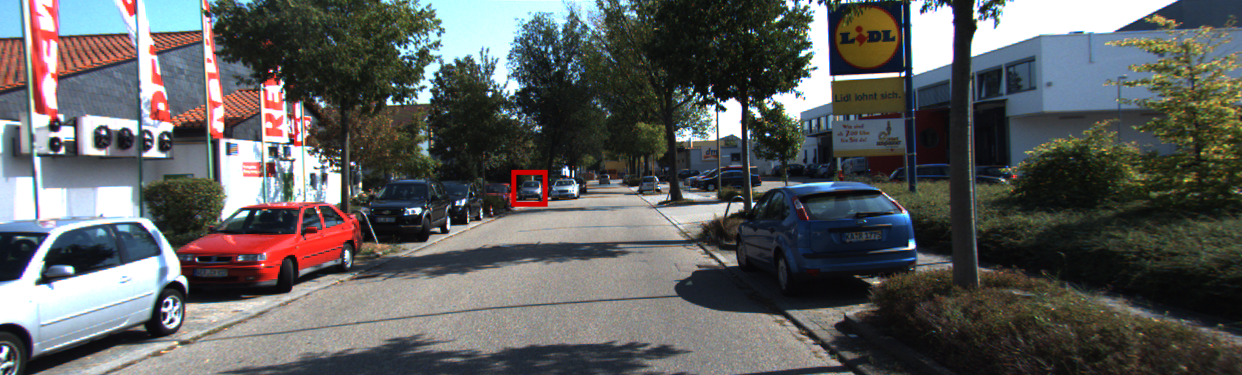}
    \caption{Proposed annotation errors in nuScenes (top two) and KITTI (bottom two).
    Top images show point clouds with annotations in purple and the proposal in red.
    Camera images aid the evaluation.
    }\label{fig:label-error-proposals}
\end{figure}
The task of annotations error detection with LMD is inspired by \cref{fig: scatter plot}.
There are a number of predictions with $\ioubev=0$ but with high predicted $\ioubev$.
After looking at these FPs it has been noticed that the prediction itself is, in fact, correct and the corresponding ground truth is not.
More precisely, incorrect ground truth corresponds to missing labels, labels with a wrong assigned class or the location of the annotation is inaccurate, i.e., the 3D bounding box is not correctly aligned with the point cloud.
Annotation error detection with LMD works as follows: all FP predictions, i.e., predictions that have $\ioubev<0.5$ with the ground truth, are sorted by the predicted $\ioubev$ in a descending order across all images.
Then, the first 100 predictions, i.e., the top 100 FPs with highest predicted $\ioubev$, are manually reviewed, see \cref{fig:label-error-proposals} for examples of proposals by this method.
In this case, a GB classifier is used to predict the box-wise $\ioubev$.
We compare LMD against a score baseline which works in the same way, except that the FPs are sorted by the objectness score.
As a random baseline, 100 randomly drawn FPs are considered for review which provides an insight into how well the respective test dataset is labeled.
In general, if it was unclear whether an annotation error was present or not, this case was not marked as annotation error, i.e., the following numbers are a conservative (under-)estimation.
LMD finds 43 annotation errors from 100 proposals and, in contrast, the score only 6 out of 100.
Even the random baseline still finds 3 annotation errors, which indicates that there is a significant number of annotation errors in the nuScenes test dataset and that these can be found at far smaller effort with LMD than with the score.
Annotation error detection counts for nuScenes and KITTI test datasets are shown in the appendix.

\section{Conclusion}
In this work we have introduced a purely post-processing-based uncertainty quantification method (LMD).
A post-processing module, which is simple to fit and can be plugged onto any pre-trained Lidar object detector, allows for swift estimation of confidence (meta classification) and localization precision (meta regression) in terms of $\ioubev$ at inference time.
Our experiments show that separation of true and false predictions obtained from LMD is sharper than that of the base detector.
Statistical reliability is significantly improved in terms of calibration of the obtained confidence scores and $\ioubev$ is estimated to considerable precision at inference time, i.e., without knowledge of the ground truth.
In addition to statistical improvement in decision making, we introduce a method for detecting annotation errors in real-world datasets based on our uncertainty estimation method.
Error counts of hand-reviewed proposals which are shown for broadly used public benchmark datasets suggest a highly beneficial industrial use case of our method beyond improving prediction reliability.
We also hope that our investigations will spark future research in the domains of light-weight uncertainty estimation and annotation error detection for large-scale datasets. 

\section*{Acknowledgement}
We gratefully acknowledge financial support by the state Ministry of Economy, Innovation and Energy of Northrhine Westphalia (MWIDE) and the European Fund for Regional Development via the FIS.NRW project BIT, grant no.\ EFRE-0400216, as well as ``Projekt UnrEAL'', grant no.\ 01IS22069, funded by the German Federal Ministry of Education and Research.

%
%
%
\bibliographystyle{splncs04}
\bibliography{egbib.bib}

\newpage
\appendix
\section{Correlation of Box-wise Features with the $\ioubev$}
\begin{table}
    \centering
    
    \caption{Strongest correlation coefficients for constructed box-wise features and $\ioubev$ for the CenterPoint (left) and PointPillars (right) architecture on the KITTI test dataset and a score threshold $\tau$ = 0.1. \hfill}
    
    \resizebox{.48\columnwidth}{!}{
    \begin{tabular}{lc|lc}
        \toprule
        $\max\{s^i\}$             & 0.8056 & $s^i$                               & 0.8050 \\ 
        $\mathrm{std}\{s^i\}$              & 0.7456 & $\mathrm{mean}\{s^i\}$            & 0.7289 \\ 
        $\mathrm{mean}\{\rho^i_{\max}\}$          & 0.5769 & $\max\{\rho^i_{\max}\}$               & 0.5768 \\ 
        $\rho^i_{\max}$          & 0.5739 & $\min\{\rho^i_{\max}\}$          & 0.5381 \\ 
        $\mathrm{mean}\{\ioubev^i\}$ & 0.5083 & $\min\{\iouthreed^i\}$  & 0.4984 \\ 
        $\max\{\iouthreed^i\}$   & 0.4974 & $\max\{P^i\}$                      & 0.4593 \\ \bottomrule
    \end{tabular}
    }
    \hfill
    \resizebox{.48\columnwidth}{!}{
    \begin{tabular}{lc|lc}
        \toprule
        $\max\{s^i\}$ & 0.8493 & $s^i$ & 0.8490 \\ 
        $\mathrm{mean}\{s^i\}$\phantom{$\{IO\}$} & 0.8309 & $\mathrm{std}\{s^i\}$\phantom{$\{IO\}$} & 0.7116 \\ 
        $N^i$ & 0.6258 & $\max\{\ell^i\}$ & 0.5631 \\ 
        $\max\{w^i\}$ & 0.5174 & $\mathrm{mean}\{\ell^i\}$ & 0.5159 \\ 
        $\max\{\rho^i_{\max}\}$ & 0.5144 & $\mathrm{mean}\{w^i\}$ & 0.5129 \\ 
        $\max\{F^i\}$ & 0.5120 & $\mathrm{mean}\{F^i\}$ & 0.5089 \\ \bottomrule
    \end{tabular}
    }
    \label{tab: kitti_highest_corr} 
\end{table}
\begin{table}
    \centering
    
    \caption{Strongest correlation coefficients for constructed box-wise features and $\ioubev$ for the CenterPoint (left) and PointPillars (right) architecture on the nuScenes test dataset and a score threshold $\tau$ = 0.1. \hfill}
    
    \resizebox{.48\columnwidth}{!}{
    \begin{tabular}{lc|lc}
        \toprule
        $\max\{s^{i}\}$   & 0.7516 & $\mathrm{std}\{s^{i}\}$     & 0.6991 \\ 
        $s^{i}$                      & 0.6755 & $\mathrm{mean}\{s^{i}\}$    & 0.5847 \\ 
        $N^{i}$                      & 0.3007 & $\max\{\iouthreed^i\}$   & 0.2900 \\ 
        $\mathrm{mean}\{\iouthreed^{i}\}$ & 0.2707 & $\mathrm{std}\{\rho^{i}_\mathrm{max}\}$  & 0.2652 \\ 
        $\mathrm{std}\{\iouthreed^{i}\}$  & 0.2560 & $\max\{\rho^{i}_\mathrm{max}\}$ & 0.2556 \\ 
        $\max\{P^{i}\}$       & 0.2519 & $\mathrm{mean}\{P^{i}\}$       & 0.2500 \\ \bottomrule
    \end{tabular}
    }
    \hfill
    \resizebox{.48\columnwidth}{!}{
    \begin{tabular}{lc|lc}
        \toprule
        $\max\{s^i\}$   & 0.7554 & $\mathrm{std}\{s^i\}$   & 0.7344 \\ 
        $\mathrm{mean}\{s^i\}$   & 0.7161 & $N^i$                     & 0.6629 \\ 
        $s^i$                      & 0.6244 & $\mathrm{std}\{w^i\}$        & 0.4535 \\ 
        $\mathrm{std}\{F^i\}$    & 0.4228 & $\mathrm{std}\{y^i\}$        & 0.3879 \\ 
        $\mathrm{std}\{\ell^i\}$         & 0.3794 & $\mathrm{std}\{\iouthreed^i\}$ & 0.3689 \\ 
        $\mathrm{std}\{\ioubev^i\}$ & 0.3581 & $\mathrm{std}\{P^i\}$      & 0.3433 \\ 
        \bottomrule
    \end{tabular}
    }
    \label{tab: nuscenes_highest_corr} 
\end{table}
\begin{table}
    \centering
    
    \caption{Strongest correlation coefficients for constructed box-wise features and $\ioubev$ for the CenterPoint (left) and PointPillars (right) architecture on the \aptiv\, test dataset and a score threshold $\tau$ = 0.1. \hfill}
    
    \resizebox{.48\columnwidth}{!}{
    \begin{tabular}{lc|lc}
        \toprule
        $\max\{s^i\}$    & 0.7649 & $s^i$                       & 0.7358 \\ 
        $\mathrm{std}\{s^i\}$     & 0.6598 & $\mathrm{mean}\{s^i\}$    & 0.5913 \\ 
        $\max\{\iouthreed^i\}$   & 0.5656 & $\mathrm{mean}\{\iouthreed^i\}$  & 0.5596 \\ 
        $\max\{\ioubev^i\}$  & 0.5573 & $\mathrm{mean}\{\ioubev^i\}$ & 0.5555 \\ 
        $\ell^i$                       & 0.3736 & $\mathrm{min}\{\ell^i\}$          & 0.3731 \\ 
        $\mathrm{mean}\{\ell^i\}$         & 0.3724 & $\max\{\ell^i\}$          & 0.3694 \\ \bottomrule
    \end{tabular}
    }
    \hfill
    \resizebox{.48\columnwidth}{!}{
    \begin{tabular}{lc|lc}
        \toprule
        $\max\{s^i\}$  & 0.7596 & $\mathrm{mean}\{s^i\}$   & 0.7577 \\ 
        $\mathrm{std}\{s^i\}$  & 0.7570 & $s^i$                      & 0.7108 \\ 
        $N^i$                     & 0.5226 & $\mathrm{std}\{\ell^i\}$         & 0.5225 \\ 
        $\mathrm{std}\{F^i\}$   & 0.4488 & $\mathrm{std}\{h^i\}$         & 0.3998 \\ 
        $\mathrm{std}\{\iouthreed^i\}$\phantom{III} & 0.3879 & $\mathrm{std}\{\rho^i_{\max}\}$ & 0.3802 \\ 
        $\mathrm{std}\{z^i\}$        & 0.3727 & $\max\{\iouthreed^i\}$\phantom{III} & 0.3443 \\ \bottomrule
    \end{tabular}
    }
    \label{tab: aptiv_highest_corr} 
\end{table}

\Cref{tab: kitti_highest_corr}, \cref{tab: nuscenes_highest_corr} and \cref{tab: aptiv_highest_corr} show the Pearson correlation coefficients of the constructed box-wise dispersion measures with the $\ioubev$ of prediction and ground truth for the KITTI, nuScenes and \aptiv\, test datasets. 
Comparable to the results from \cref{fig:pearson-correlation-with-ioubev} of the main paper, the score features have strong correlations ($> 0.5$) with the $\ioubev$, independently of the underlying dataset or architecture.
Especially for the PointPillars architecture, the number of proposal boxes $N^i$ has a correlations $> 0.6$ for nuScenes and KITTI and $> 0.5$ for \aptiv, whereas for CenterPoint, $N^i$ shows minor correlations ($<0.45$).
Moreover, the overlaps of the proposal boxes (different $\iou$ features), as well as the localization of the box (especially $\ell$, $h$ and $w$) and the maximal reflectance value of points within the box ($\rho^i_{\max}$) seem to be reasonably correlated with the $\ioubev$.
Although the other features have rather smaller correlations with the $\ioubev$, they may still contribute to a more informative set of features for meta classification and regression.

\section{Comparison of Different Meta Classifiers}
\begin{table*}
    \centering
    
    \caption{Comparison of meta classification accuracy and $\auroc$ for the score baseline, bounding box features and LMD for all available network-dataset combinations with $\ioubev$ threshold $0.5$ and score threshold $0.1$. Models used for meta classification are Logistic Regression (LogReg), Gradient Boosting (GB), Random Forest (RF) and a Multi Layer Perceptron (MLP). Higher
    values are better. Bold numbers indicate the highest performance and underlined numbers represent the
    second highest (row-wise).}
    
    \resizebox{\textwidth}{!}{
    \begin{tabular}{c c c | c c c c | c c c c}
        \toprule
          &  &  & \multicolumn{4}{|c}{Accuracies} & \multicolumn{4}{|c}{$\auroc$s} \\
         \midrule
         Dataset & Network & Method & LogReg & RF & GB & MLP & LogReg & RF & GB & MLP \\
         \midrule
         \multirow{6}{*}{KITTI} & \multirow{3}{*}{PointPillars} & Score & \textbf{0.8955} & 0.8848 & 0.8921 & \underline{0.8940} & \textbf{0.9566} & 0.9497 & 0.9530 & \textbf{0.9566} \\ 
         & & Box Features & \textbf{0.8961} & \underline{0.8958} & 0.8931 & 0.8846 & \textbf{0.9564} & \underline{0.9548} & 0.9537 & 0.9482 \\ 
         & & LMD & 0.9000 & \textbf{0.9028} & \underline{0.9004} & 0.8844 & 0.9589 & \textbf{0.9621} & \underline{0.9592} & 0.9479 \\
        \cline{2-11}
        & \multirow{3}{*}{CenterPoint} & Score & \textbf{0.8726} & 0.8651 & 0.8688 & \underline{0.8725} & \textbf{0.9322} & 0.9242 & 0.9274 & \textbf{0.9322} \\ 
         & & Box Features & \textbf{0.8727} & \underline{0.8719} & 0.8691 & 0.8608 & 0.9244 & \textbf{0.9387} & \underline{0.9343} & 0.9271 \\ 
         & & LMD & \underline{0.8818} & \textbf{0.8847} & 0.8806 & 0.8700 & 0.9421 & \textbf{0.9522} & \underline{0.9466} & 0.9362 \\
         \midrule
        \multirow{6}{*}{nuScenes} & \multirow{3}{*}{PointPillars} & Score & \underline{0.8402} & 0.8106 & 0.8398 & \textbf{0.8403} & \textbf{0.8151} & 0.8130 & 0.8129 & \underline{0.8150} \\ 
         & & Box Features & 0.8409 & 0.8613 & \textbf{0.8708} & \underline{0.8653} & 0.8499 & \textbf{0.9018} & \underline{0.9002} & 0.8957 \\
         & & LMD & 0.8875 & 0.8842 & \textbf{0.8915} & \underline{0.8908} & 0.9208 & \underline{0.9257} & \textbf{0.9280} & 0.9252 \\
         \cline{2-11}
        & \multirow{3}{*}{CenterPoint} & Score & \textbf{0.8777} & 0.8524 & 0.8772 & \underline{0.8773} & \textbf{0.8644} & 0.8617 & 0.8623 & \underline{0.8640} \\ 
         & & Box Features & 0.8877 & \underline{0.9049} & \textbf{0.9203} & 0.8975 & 0.9056 & \underline{0.9454} & \textbf{0.9529} & 0.9293 \\ 
         & & LMD & 0.9118 & 0.9166 & \textbf{0.9297} & \underline{0.9200} & 0.9450 & \underline{0.9581} & \textbf{0.9628} & 0.9530 \\  
         \midrule
        \multirow{6}{*}{\aptiv} & \multirow{3}{*}{PointPillars} & Score & \textbf{0.7956} & 0.7929 & 0.7939 & \underline{0.7954} & \textbf{0.8582} & 0.8555 & 0.8558 & \underline{0.8580} \\ 
         & & Box Features & 0.8184 & \textbf{0.8508} & \underline{0.8489} & 0.8472 & 0.8933 & \textbf{0.9289} & \underline{0.9274} & 0.9255 \\ 
         & & LMD & 0.8506 & \underline{0.8599} & \textbf{0.8615} & 0.8515 & 0.9273 & \underline{0.9382} & \textbf{0.9396} & 0.9300  \\
         \cline{2-11}
        & \multirow{3}{*}{CenterPoint} & Score & \textbf{0.8279} & 0.8149 & 0.8265 & \textbf{0.8279} & \textbf{0.8946} & 0.8900 & 0.8914 & \textbf{0.8946} \\ 
         & & Box Features & 0.8340 & \textbf{0.8452} & \underline{0.8440} & 0.8439 & 0.9029 & \textbf{0.9155} & \underline{0.9134} & 0.9085 \\ 
         & & LMD & 0.8478 & \textbf{0.8559} & \underline{0.8548} & 0.8529 & 0.9187 & \textbf{0.9294} & \underline{0.9275} & 0.9227 \\
        \bottomrule
    \end{tabular}
    }
    \label{tab: meta classification comparison appendix}
\end{table*}

\Cref{tab: meta classification comparison appendix} presents meta classification accuracy and $\auroc$ for different meta classification models for the KITTI, nuScenes and \aptiv\, test datasets.
For the score baseline and besides of PointPillars and nuScenes, the linear model (LogReg) outperforms the random forest, gradient boosting and the MLP, where the difference is at most $2.96$ percentage points (pp) in terms of meta classification accuracy.
In terms of meta classification $\auroc$, the linear model outperforms all other meta models of at most $0.8$ pp.
For the bounding box features and LMD, random forest or gradient boosting are in most cases the best meta models in terms of meta classification accuracy and $\auroc$.
In general, the bounding box features outperform the score baseline and LMD outperforms the bounding box features.

\section{Comparison of Different Meta Regressors}
\begin{table*}
    \centering
    \caption{Comparison of meta regression $R^2$ for the score baseline, bounding box features and LMD for all available network-dataset combinations using an $\ioubev$ threshold of $0.5$ and score threshold of $0.1$. Models used for meta regression are Ridge Regression (RR), Gradient Boosting (GB), Random Forest (RF) and a Multi Layer Perceptron (MLP). Higher
    values are better. Bold numbers indicate the highest performance and underlined numbers represent the
    second highest (row-wise).}
    \resizebox{.65\textwidth}{!}{
    \begin{tabular}{c c c | c c c c}
        \toprule
          &  &  & \multicolumn{4}{|c}{$R^2$} \\
         \midrule
         Dataset & Network & Method & RR & RF & GB & MLP \\
         \midrule
         \multirow{6}{*}{KITTI} & \multirow{3}{*}{PointPillars} & Score & 0.6901 & 0.6726 & \underline{0.7108} & \textbf{0.7146} \\ 
         & & Box Features & 0.6973 & \underline{0.7044} & \textbf{0.7131} & 0.6819 \\ 
         & & LMD & 0.7151 & \textbf{0.7301} & \underline{0.7287} & 0.6837 \\
        \cline{2-7}
        & \multirow{3}{*}{CenterPoint} & Score & 0.6220 & 0.5877 & \underline{0.6235} & \textbf{0.6273}  \\ 
         & & Box Features & 0.6260 & \underline{0.6446} & \textbf{0.6472} & 0.6274 \\ 
         & & LMD & 0.6631 & \textbf{0.6863} & \underline{0.6840} & 0.6538 \\
         \midrule
        \multirow{6}{*}{nuScenes} & \multirow{3}{*}{PointPillars} & Score & 0.3903 & 0.4006 & \textbf{0.4055} & \underline{0.4054} \\  
         & & Box Features & 0.4187 & \underline{0.5586} & \textbf{0.5593} & 0.5356 \\
         & & LMD & 0.6105 & \underline{0.6346} & \textbf{0.6413} & 0.6244 \\
         \cline{2-7}
        & \multirow{3}{*}{CenterPoint} & Score & 0.4641 & 0.4675 & \underline{0.4733} & \textbf{0.4751} \\ 
         & & Box Features & 0.5292 & \underline{0.6681} & \textbf{0.6792} & 0.6249 \\ 
         & & LMD & 0.6451 & \underline{0.7242} & \textbf{0.7296} & 0.7122 \\  
         \midrule
        \multirow{6}{*}{\aptiv} & \multirow{3}{*}{PointPillars} & Score & 0.5005 & 0.5013 & \underline{0.5096} & \textbf{0.5106} \\ 
         & & Box Features & 0.5469 & \underline{0.6484} & \textbf{0.6568} & 0.6482 \\ 
         & & LMD & 0.6401 & \underline{0.6830} & \textbf{0.6924} & 0.6614 \\
         \cline{2-7}
        & \multirow{3}{*}{CenterPoint} & Score & \underline{0.5458} & 0.5329 & 0.5456 & \textbf{0.5469} \\ 
         & & Box Features & 0.5749 & \underline{0.6200} & \textbf{0.6286} & 0.6136 \\ 
         & & LMD & 0.6210 & \underline{0.6541} & \textbf{0.6591} & 0.6332 \\
        \bottomrule
    \end{tabular}
}
    \label{tab: meta regression comparison appendix}
\end{table*}

\Cref{tab: meta regression comparison appendix} states meta regression $R^2$ for different meta regression models for the KITTI, nuScenes and \aptiv\, test datasets.
Random forest and gradient boosting outperforms the linear model (ridge regression) and the MLP in every case for the bounding box features and LMD.
Except for CenterPoint on \aptiv, both MLP and gradient boosting are the superior meta models (compared to random forest and ridge regression) in terms of meta regression $R^2$ for the score baseline. 
Comparable to the results of \cref{tab: meta model comparison} of the main paper and the results for meta classification (\cref{tab: meta classification comparison appendix}), the bounding box features outperform the score baseline and LMD outperforms the bounding box features.

\section{Feature Selection for Meta Classification and Meta Regression}
\begin{table*}
    \centering
    \caption{Feature selection for meta classification $\auroc$ using a greedy heuristic for all network-dataset combinations, score threshold $0.1$ and a GB classifier. The right-most column shows the the performance when incorporating all features (LMD).}
    \resizebox{.95\textwidth}{!}{
    \begin{tabular}{c c || c | c | c | c | c | c | c | c | c | c || c}
    \toprule
        \multicolumn{13}{c}{Meta Classification}  \\
        \toprule
       &  &  \multicolumn{11}{|c}{Number of Features} \\
      \toprule
      Network & Meta Model & 1 & 2 & 3 & 4 & 5 & 6 & 7 & 8 & 9 & 10 & All \\
      \toprule
      \multirow{2}{*}{KITTI} & PointPillars & 0.9482 & 0.9512 & 0.9530 & 0.9543 & 0.9564 & 0.9576 & 0.9580 & 0.9584 & 0.9587 & 0.9588 & 0.9592 \\
       & CenterPoint & 0.9149 & 0.9268 & 0.9372 & 0.9405 & 0.9420 & 0.9439 & 0.9452 & 0.9458 & 0.9462 & 0.9463 & 0.9466 \\
       \midrule
       \multirow{2}{*}{nuScenes} & PointPillars & 0.9117 & 0.9175 & 0.9214 & 0.9248 & 0.9253 & 0.9257 & 0.9259 & 0.9263 & 0.9268 & 0.9273 & 0.9280 \\
       & CenterPoint & 0.9253 & 0.9417 & 0.9591 & 0.9603 & 0.9609 & 0.9611 & 0.9614 & 0.9617 & 0.9619 & 0.9620 & 0.9628 \\
       \midrule
       \multirow{2}{*}{\aptiv} & PointPillars & 0.8940 & 0.9182 & 0.9233 & 0.9288 & 0.9312 & 0.9330 & 0.9339 & 0.9348 & 0.9352 & 0.9356 & 0.9396 \\
       & CenterPoint & 0.9051 & 0.9150 & 0.9177 & 0.9195 & 0.9209 & 0.9226 & 0.9242 & 0.9255 & 0.9259 & 0.9268 & 0.9275 \\
       \bottomrule
    \end{tabular}
    }
    \label{tab: greedy meta classification appendix}
\end{table*}
\begin{table*}
    \centering
    \onehalfspacing
    \caption{Feature selection for meta regression $R^2$ using a greedy heuristic for all network-dataset combinations, score threshold $0.1$ and a GB regressor. The right-most column shows the the performance when incorporating all features (LMD).}
    \resizebox{.95\textwidth}{!}{
    \begin{tabular}{c c || c | c | c | c | c | c | c | c | c | c || c}
    \toprule
      \multicolumn{13}{c}{Meta Regression}  \\
      \toprule
       &  &  \multicolumn{11}{|c}{Number of Features} \\
      \toprule
      Network & Meta Model & 1 & 2 & 3 & 4 & 5 & 6 & 7 & 8 & 9 & 10 & All \\
      \toprule
      \multirow{2}{*}{KITTI} & PointPillars & 0.7053 & 0.7121 & 0.7165 & 0.7195 & 0.7217 & 0.7229 & 0.7237 & 0.7248 & 0.7260 & 0.7267 & 0.7287 \\
       & CenterPoint & 0.6134 & 0.6449 & 0.6626 & 0.6746 & 0.6776 & 0.6791 & 0.6808 & 0.6815 & 0.6824 & 0.6833 & 0.6840 \\
       \midrule
       \multirow{2}{*}{nuScenes} & PointPillars & 0.5931 & 0.6069 & 0.6174 & 0.6265 & 0.6318 & 0.6359 & 0.6383 & 0.6390 & 0.6397 & 0.6401 & 0.6413 \\
       & CenterPoint & 0.5892 & 0.6591 & 0.7079 & 0.7158 & 0.7204 & 0.7243 & 0.7264 & 0.7280 & 0.7286 & 0.7289 & 0.7296 \\
       \midrule
       \multirow{2}{*}{\aptiv} & PointPillars & 0.5860 & 0.6300 & 0.6454 & 0.6615 & 0.6679 & 0.6740 & 0.6774 & 0.6806 & 0.6845 & 0.6875 & 0.6924 \\
       & CenterPoint & 0.5856 & 0.6301 & 0.6358 & 0.6400 & 0.6472 & 0.6500 & 0.6537 & 0.6558 & 0.6564 & 0.6572 & 0.6591 \\
       \bottomrule
    \end{tabular}
    }
    \label{tab: greedy meta regression appendix}
\end{table*}
\Cref{tab: greedy meta classification appendix} shows feature selection results in terms of meta classification $\auroc$ and \cref{tab: greedy meta regression appendix} shows feature selection results in terms of meta regression $R^2$. 
For both tasks, meta classification and regression, a few features are sufficient to reach roughly the same performance as when using all features (LMD). 
Meta models using five or more selected features are at most 
$0.84$ pp below the performance of LMD in terms of meta classification $\auroc$ and at most $2.45$ pp in terms of meta regression $R^2$.
With ten features used, the respective differences to the performance results achieved by LMD are $\leq 0.4$ pp in terms of meta classification $\auroc$ and $<0.5$ pp in terms of meta regression $R^2$.

\section{Annotation Error Detection}
\begin{table}
\centering
\onehalfspacing
\caption{Comparison of detected annotation errors for the KITTI test dataset using object detectors as baselines and RF as best LMD classifier with a score threshold of $0.1$ and an $\ioubev$ threshold of $0.5$.}
\resizebox{.5\textwidth}{!}{
\begin{tabular}{cc|ccc}
\toprule
\multicolumn{5}{c}{KITTI Annotation Error Analysis (RF)} \\ 
\toprule
Network & Classes & Random & Smart & LMD \\ 
\toprule
\multirow{4}{*}{PointPillars} & Pedestrian & $8/53$ & $1/1$ & $2/4$ \\ 
\cline{2-5}
 & Cyclist & $3/22$ & $1/1$ & $2/2$ \\ \cline{2-5}
 & Car & $9/25$ & $76/98$ & $88/94$ \\ \cline{2-5}
 & Overall & \textbf{20/100} & \textbf{78/100} & \textbf{92/100} \\ 
\midrule
\multirow{4}{*}{CenterPoint} & Pedestrian & $4/25$ & $25/40$ & $7/7$ \\ 
\cline{2-5}
 & Cyclist & $1/11$ & $8/12$ & $1/1$ \\
 \cline{2-5}
 & Car & $22/64$ & $38/48$ & $89/92$ \\ 
 \cline{2-5}
 & \textbf{Overall} & \textbf{27/100} & \textbf{71/100} & \textbf{97/100} \\ 
\bottomrule
\end{tabular}
}
\label{tab: kitti_label_error_comparison_appendix} 
\end{table}
\begin{table*}
\centering
\caption{Comparison of detected annotation errors for the nuScenes test dataset using object detectors as baselines and GB as best LMD classifier with a score threshold of $0.1$ and an $\ioubev$ threshold of $0.5$.}
\resizebox{.55\columnwidth}{!}{
\begin{tabular}{cc|ccc}
\hline
\multicolumn{5}{c}{nuScenes Annotation Error Analysis (GB)} \\ 
\toprule
Network & Classes & Random & Score & LMD  \\ 
\toprule
\multirow{11}{*}{PointPillars} & Car & $1/27$ & $10/55$ & $13/49$ \\
\cline{2-5}
 & Pedestrian & $1/29$ & $4/10$ & $-$ \\ 
 \cline{2-5}
 & Barrier & $0/18$ & $-$ & $-$ \\ 
 \cline{2-5}
 & Traffic Cone & $0/10$ & $-$ & $-$ \\ 
 \cline{2-5}
 & Truck & $0/6$ & $13/23$ & $12/30$ \\ 
 \cline{2-5}
 & Trailer & $0/1$ & $0/10$ & $1/6$ \\ 
 \cline{2-5}
 & Bicycle & $-$ & $-$ & $-$ \\ 
 \cline{2-5}
 & Construction Vehicle & $0/4$ & $-$ & $0/6$ \\ 
 \cline{2-5}
 & Bus & $0/2$ & $0/2$ & $2/9$ \\ 
 \cline{2-5}
 & Motorcycle & $0/3$ & $-$ & $-$ \\ 
 \cline{2-5}
 & \textbf{Overall} & \textbf{2/100} & \textbf{27/100} & \textbf{28/100} \\ 
\midrule
\multirow{11}{*}{CenterPoint} & Car & $2/51$ & $0/8$ & $31/62$ \\ 
\cline{2-5}
 & Pedestrian & $-$ & $5/14$ & $2/2$ \\ 
 \cline{2-5}
 & Barrier & $0/8$ & $1/15$ & $0/1$ \\ 
 \cline{2-5}
 & Traffic Cone & $-$ & $0/4$ & $-$ \\ 
 \cline{2-5}
 & Truck & $0/14$ & $0/3$ & $8/30$ \\ 
 \cline{2-5}
 & Trailer & $0/15$ & $0/21$ & $0/1$ \\ 
 \cline{2-5}
 & Bicycle & $-$ & $-$ & $-$ \\ 
 \cline{2-5}
 & Construction Vehicle & $0/14$ & $-$ & $2/3$ \\ 
 \cline{2-5}
 & Bus & $1/8$ & $0/33$ & $0/1$ \\ 
 \cline{2-5}
 & Motorcycle & $-$ & $0/2$ & $-$ \\ 
 \cline{2-5}
 & \textbf{Overall} & \textbf{3/100} & \textbf{6/100} & \textbf{43/100} \\ 
 \bottomrule
\end{tabular}
}
\label{tab: nuscenes_label_error_comparison_appendix} 
\end{table*}
\Cref{tab: kitti_label_error_comparison_appendix} presents annotation error detection results for the KITTI test dataset.
In each experiment, we manually reviewed 100 candidates provided by a given detection method. 
For the random baseline (randomly reviewing FPs of the network) applied to PointPillars, we discover $20$ annotation errors. 
Using CenterPoint, this number increases to $27$ annotations errors with the random baseline and CenterPoint, which indicates that there might be significant number of annotation errors in the KITTI test dataset.
This is already confirmed by the smart score baseline. 
Combining it with PointPillars, we find $78$ annotation errors and with CenterPoint, we find $71$. Although these numbers seem enormous, LMD is capable of detecting even more annotation errors.
With PointPillars we detect $92$ and with CenterPoint $97$ annotation errors.

\Cref{tab: nuscenes_label_error_comparison_appendix} shows annotation error detection results for the nuScenes test dataset.
We detect only $6$ annotation errors using the smart score baseline and CenterPoint, whereas we can find $43$ annotation errors using LMD and CenterPoint.
Considering PointPillars, the gap between the detection results of the smart score baseline and LMD vanishes. 
With the smart score baseline we detect $27$ and with LMD $28$ annotation errors. 
This observation is in agreement with \cref{tab: meta classification comparison appendix}, where CenterPoint achieves superior $\auroc$ values compared to PointPillars on the nuScenes test dataset.

\section{Further Annotation Error Proposals}

\begin{figure}
    \centering
    \includegraphics[width=0.7\textwidth,trim={0 5cm 0 4cm},clip]{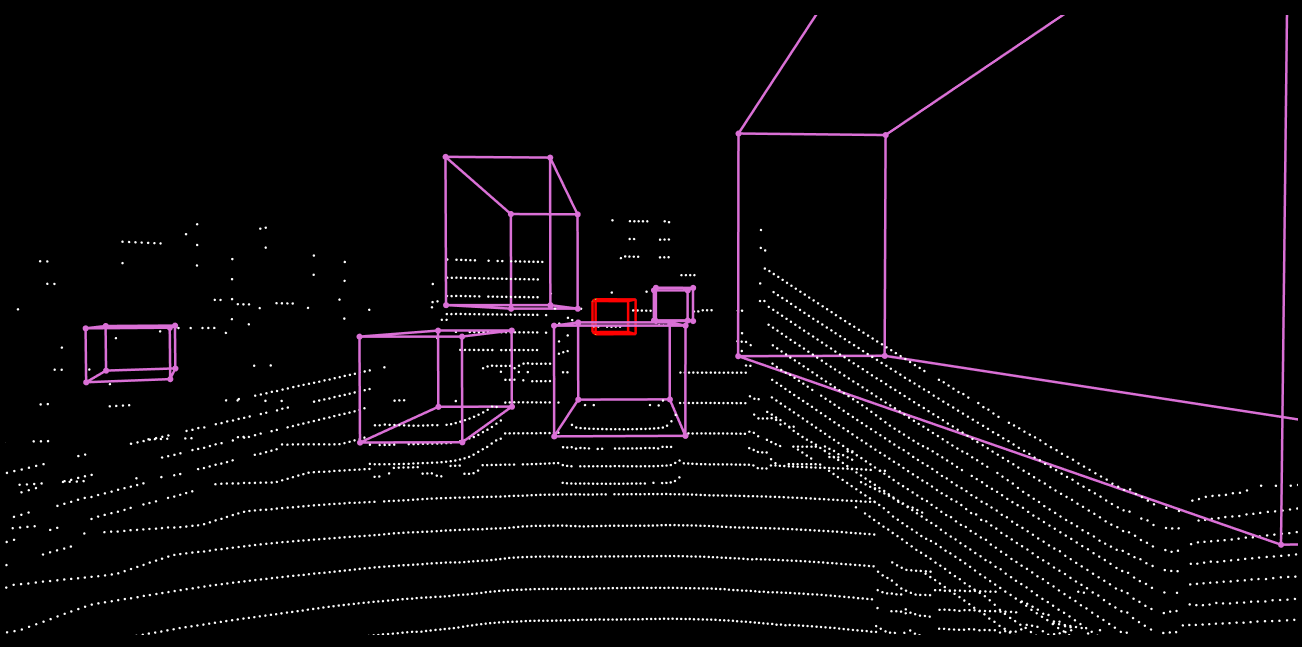}
    \includegraphics[width=0.7\textwidth,trim={0 7cm 0 8cm},clip]{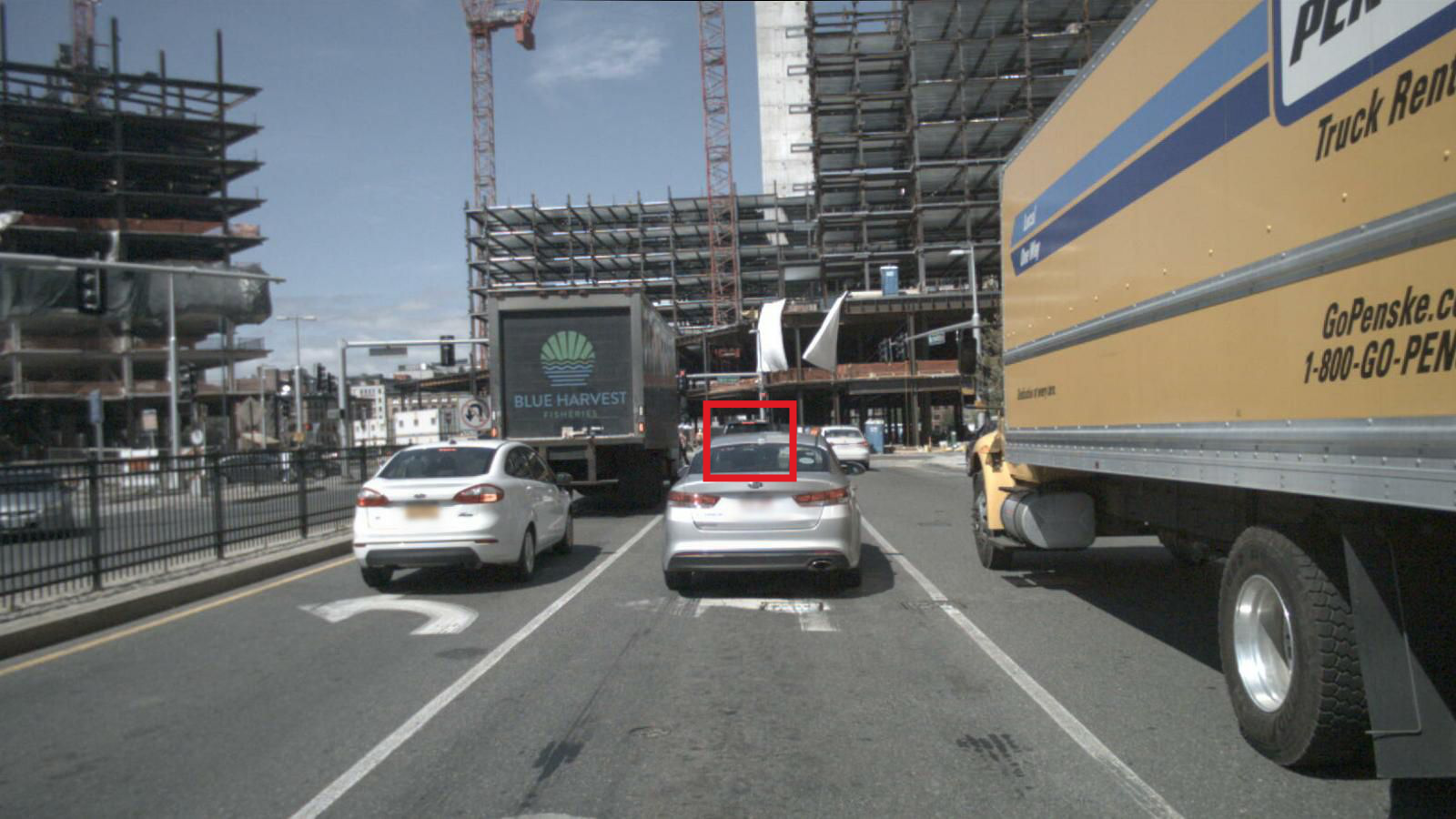}
    \includegraphics[width=0.7\textwidth,trim={0 4cm 0 2cm},clip]{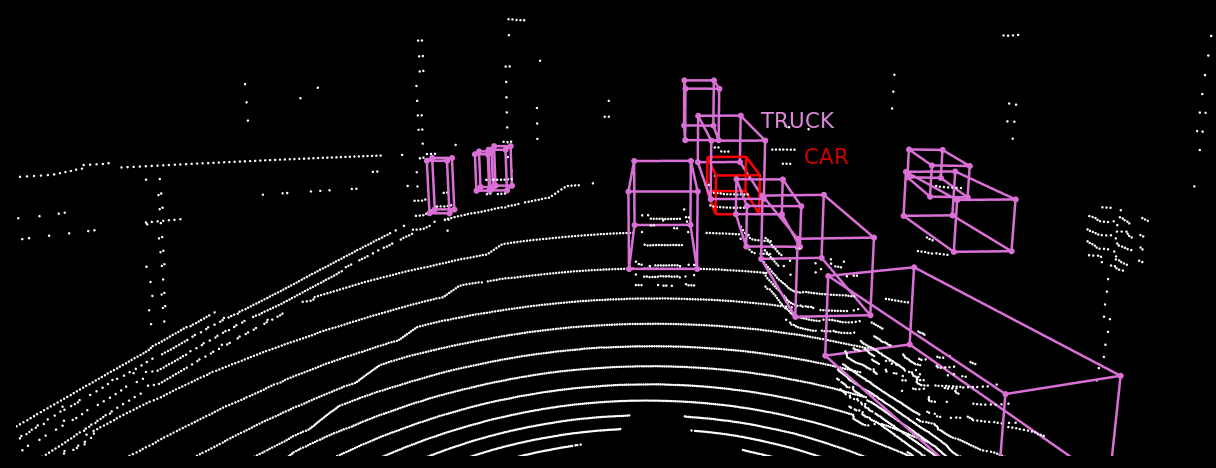}
    \includegraphics[width=0.7\textwidth,trim={0 7cm 0 9cm},clip]{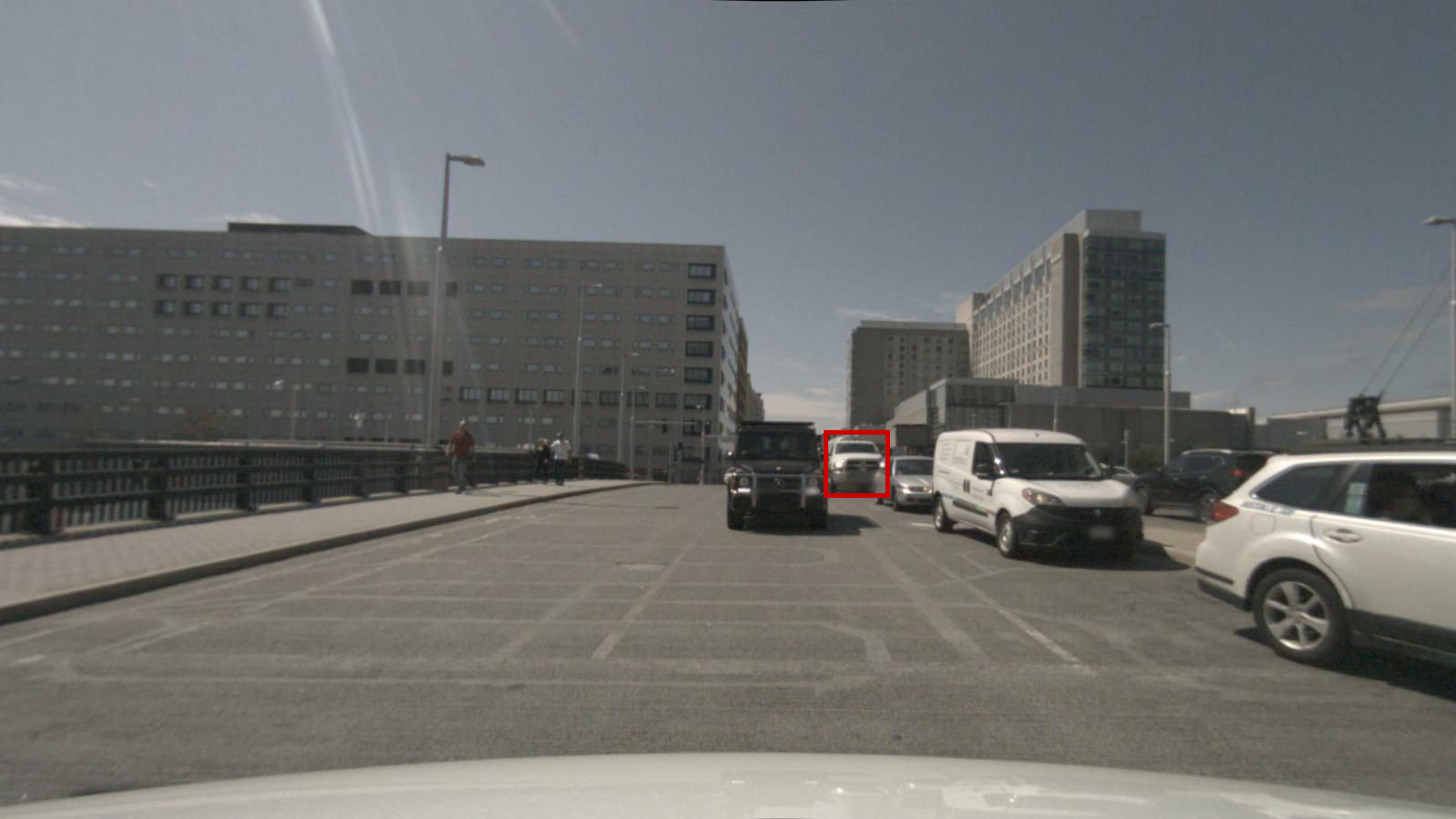}
    \includegraphics[width=0.7\textwidth,trim={0 2cm 0 5cm},clip]{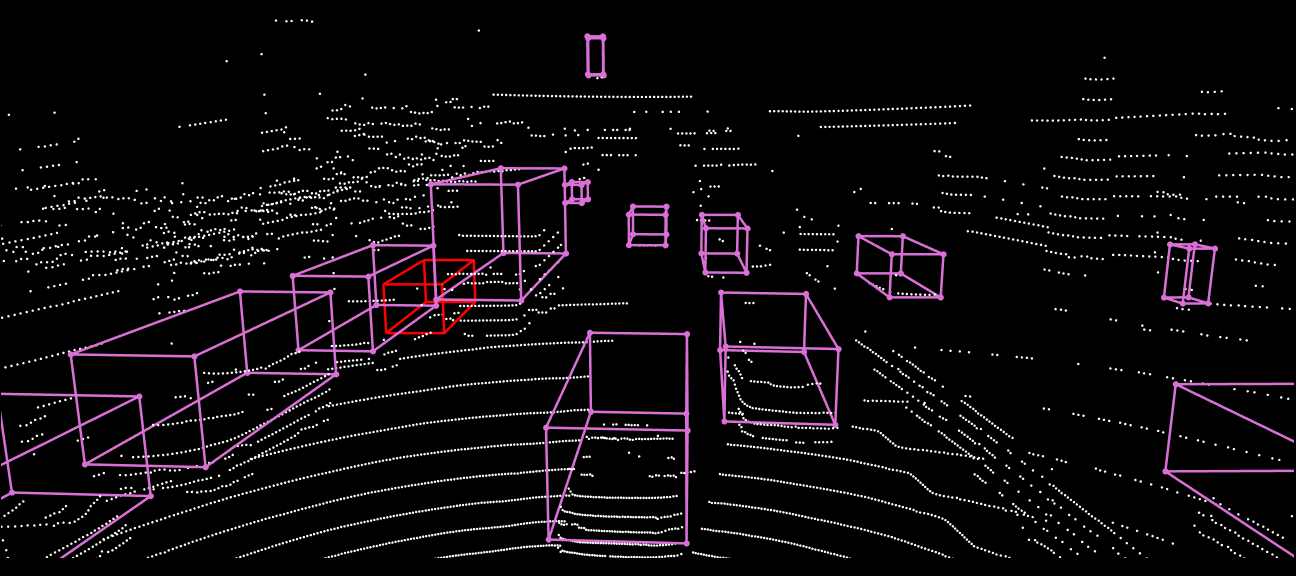}
    \includegraphics[width=0.7\textwidth,trim={0 7cm 0 8cm},clip]{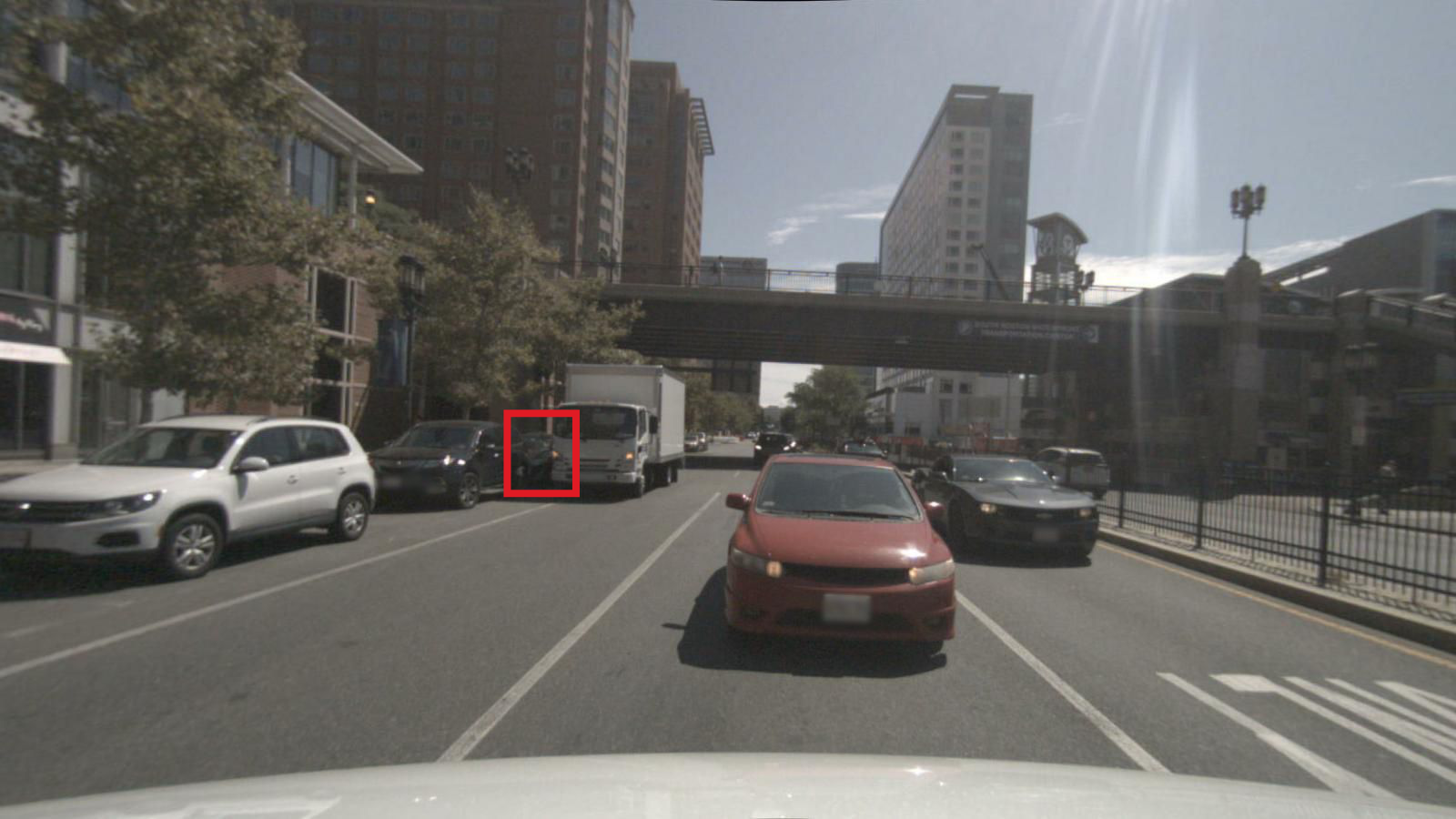}
    \caption{Additional annotation error proposals on the nuScenes test dataset.}\label{fig:additional-proposals-nuscenes}
\end{figure}

\begin{figure}
    \centering
    \includegraphics[width=0.65\textwidth,trim={0 4cm 0 4cm},clip]{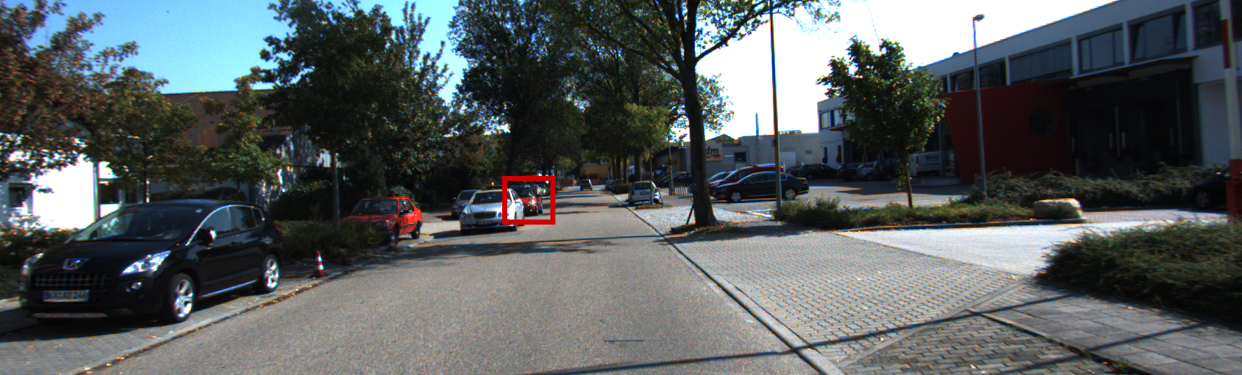}
    \includegraphics[width=0.65\textwidth,trim={0 4cm 0 2cm},clip]{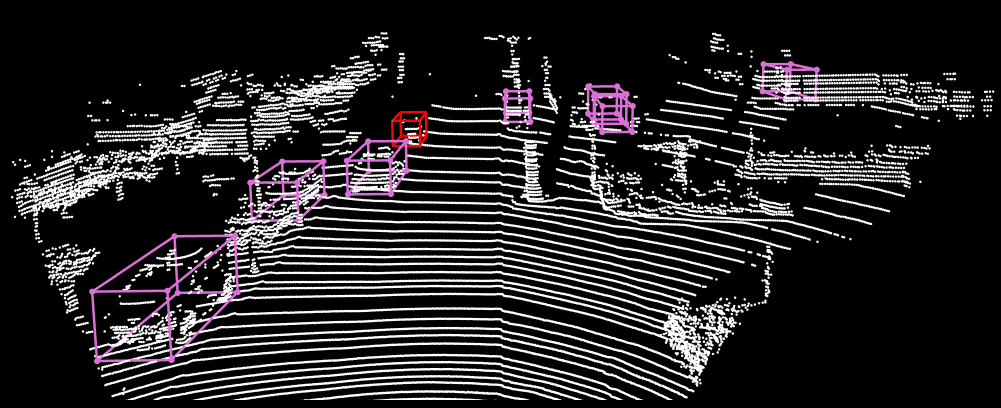}
    \includegraphics[width=0.65\textwidth,trim={0 0 0 2cm},clip]{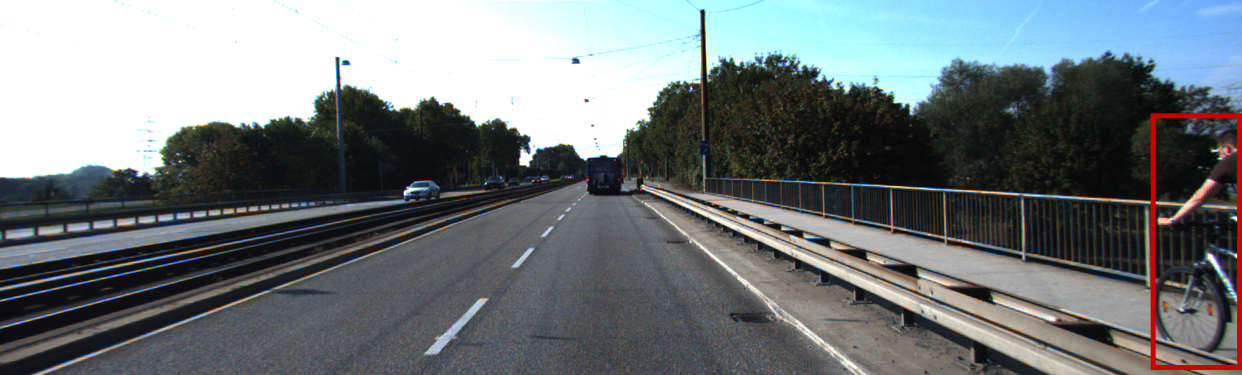}
    \includegraphics[width=0.65\textwidth,trim={0 0 0 8cm},clip]{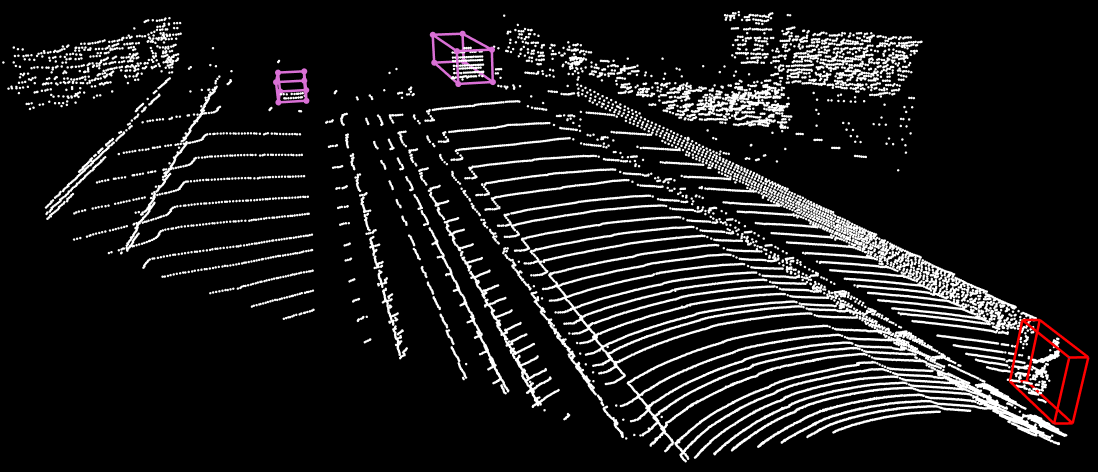}
    \includegraphics[width=0.65\textwidth,trim={0 2cm 0 4cm},clip]{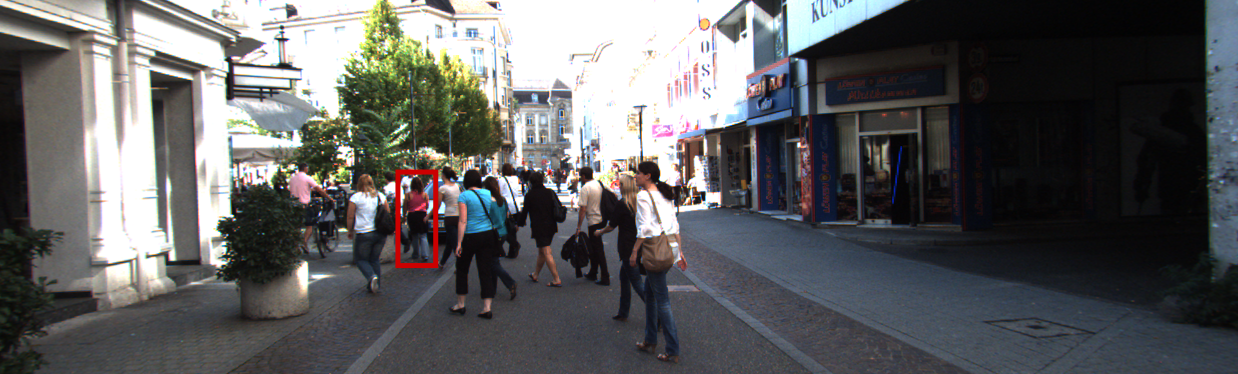}
    \includegraphics[width=0.65\textwidth,trim={0 4cm 0 8cm},clip]{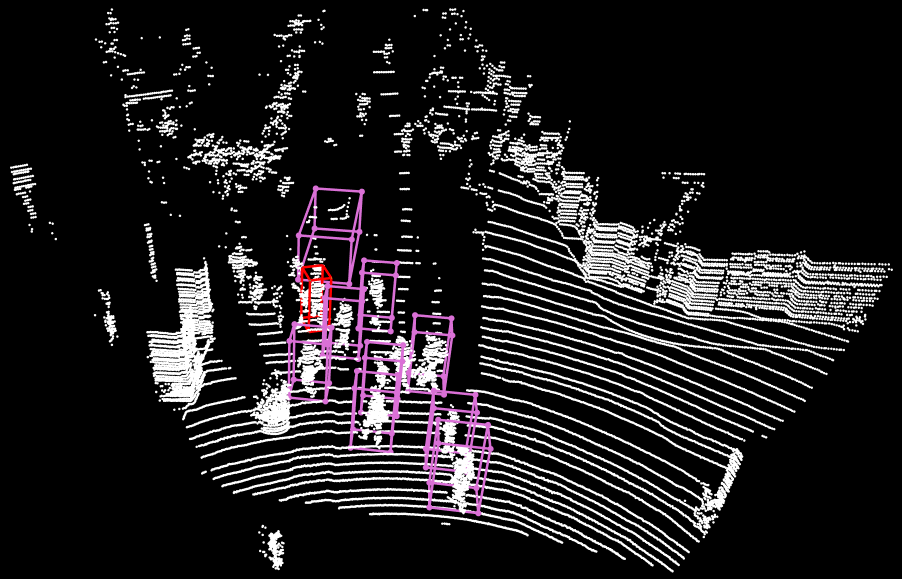}
    \includegraphics[width=0.65\textwidth,trim={0 2cm 0 4cm},clip]{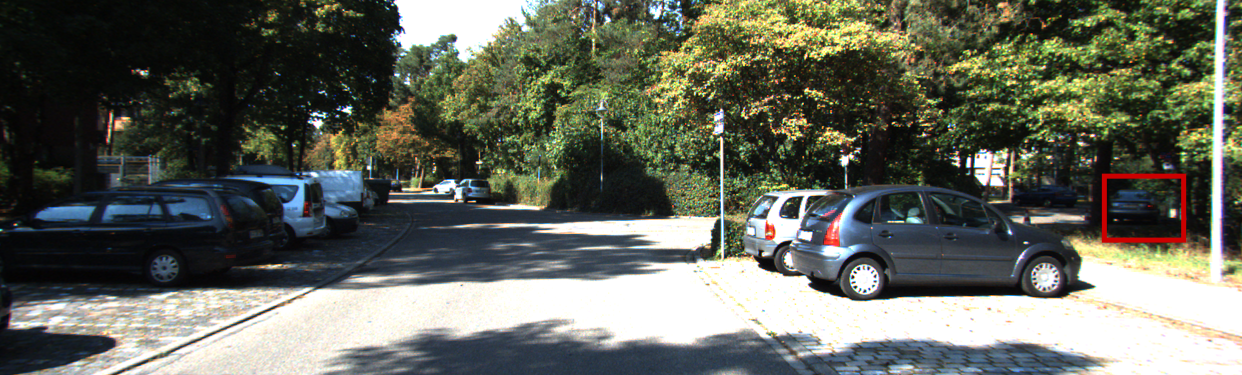}
    \includegraphics[width=0.65\textwidth,trim={0 3cm 0 4cm},clip]{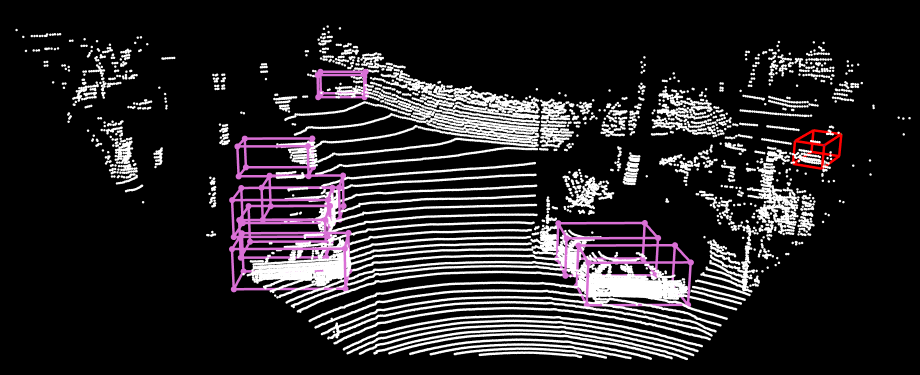}
    \includegraphics[width=0.65\textwidth,trim={0 4cm 0 4cm},clip]{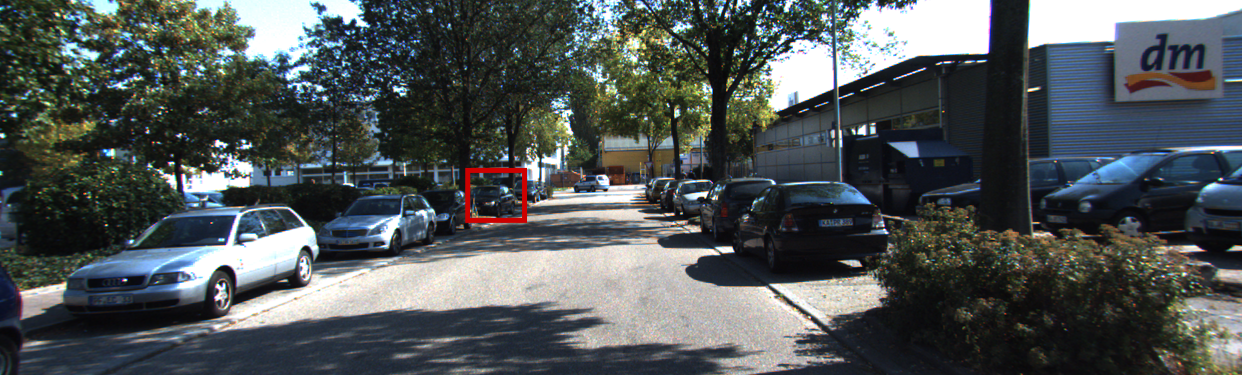}
    \includegraphics[width=0.65\textwidth,trim={0 11cm 0 2cm},clip]{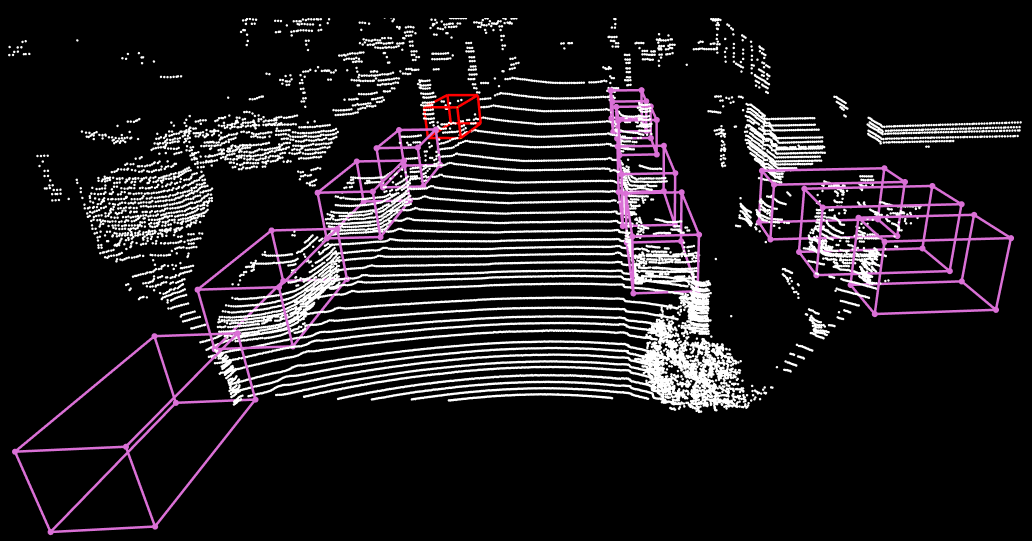}
    \caption{Additional annotation error proposals on the KITTI test dataset.}
    \label{fig:additional-proposals-kitti}
\end{figure}

Supplementing the samples of our annotation error proposal method shown in the main part of the paper, we show additional proposals for the nuScenes test dataset in \cref{fig:additional-proposals-nuscenes} and respectively for the KITTI test dataset in \cref{fig:additional-proposals-kitti}.

\end{document}